\documentclass[letterpaper]{article} 
\usepackage[preprint]{aaai2027}  
\usepackage[hyphens]{url}  
\usepackage{graphicx} 
\usepackage{natbib}  
\usepackage{caption} 
\usepackage{algorithm}
\usepackage{algorithmic}

\usepackage{newfloat}
\usepackage{listings}
\DeclareCaptionStyle{ruled}{labelfont=normalfont,labelsep=colon,strut=off} 
\floatstyle{ruled}
\newfloat{listing}{tb}{lst}{}
\floatname{listing}{Listing}

\usepackage{booktabs}

\title{Test-Time Curriculum for Open-Set AIGC Detection}
\author{%
    Yiqian Zhang\textsuperscript{\rm 1,2}\thanks{Work done during an internship at Baidu Inc.},
    Zheyuan Gu\textsuperscript{\rm 2}, 
    Xiangzhao Hao\textsuperscript{\rm 2}, 
    Zefeng Zhang,\textsuperscript{\rm 2} 
    Jingjia Mao\textsuperscript{\rm 2},\\
    Jiahao Hu\textsuperscript{\rm 2},
    Jiaxu Miao\textsuperscript{\rm 3}, 
    Jun Yu\textsuperscript{\rm 3},
    Zhenyu Zhang\textsuperscript{\rm 2}\thanks{Corresponding author.},\;
    Shuohuan Wang\textsuperscript{\rm 2},
    Yu Sun\textsuperscript{\rm 2}\\
}
\affiliations{
    \textsuperscript{\rm 1}
    Hangzhou Dianzi University \\
    \textsuperscript{\rm 2}
    ERNIE Team, Baidu Inc. \\
    \textsuperscript{\rm 3}
    Harbin Institute of Technology (Shenzhen) \\
    [1mm]
    \texttt{yiqian.zyq@gmail.com}, \; \texttt{zhangzhenyu07@baidu.com}\
}

\usepackage{enumitem}
\usepackage{microtype}
\usepackage{nicematrix}
\usepackage{makecell}
\usepackage[table]{xcolor}
\usepackage{adjustbox}
\definecolor{tableblue}{RGB}{220, 235, 250}  

\newenvironment{packed_lefty_item}{
\begin{itemize}[leftmargin=*]
\vspace{-2pt}
  \setlength{\itemsep}{0pt}
  \setlength{\parskip}{0pt}
  \setlength{\parsep}{0pt}
  \setlength{\topsep}{-10pt}
  \setlength{\partopsep}{0pt}
}{\end{itemize}\vspace{-2pt}}

\newcommand{\ie}{\textit{i}.\textit{e}.}

\begin{document}

\maketitle

\begin{abstract}
\label{sec:abstract}
AI-generated image detectors deployed in open-world environments inevitably face distribution shifts as new and stronger generative models continue to emerge. Although existing methods improve cross-generator generalization through better representations or training data construction, they typically follow a static train-once-and-deploy paradigm and cannot adapt after deployment. In this work, we study open-set AIGC image detection from a test-time adaptation perspective. We propose Test-Time Curriculum (TTC), a simple and model-agnostic framework that adapts a detector on unlabeled test data through curriculum-based self-training. TTC starts from highly reliable pseudo-labeled samples and progressively incorporates harder yet informative cases, while enforcing class-balanced selection to reduce biased updates under generator shift. To further improve pseudo-label quality, we introduce Cross-Scale Pseudo-Label Refinement, which aggregates complementary evidence across multiple resolutions for more reliable adaptation, and applies noisy-or fusion at inference to strengthen final predictions. In addition, we construct AIGCGuard, a new benchmark containing 3,100 representative real images and 124,000 generated images from 40 of the most advanced open-source and proprietary text-to-image models. Extensive experiments on five benchmarks show that TTC substantially improves overall detection performance under diverse unseen-generator shifts, establishing a practical and effective test-time adaptation framework for open-set generated image detection.
\end{abstract}
\section{Introduction}

The rapid progress of image generation models has made AI-generated images increasingly difficult to distinguish from real photographs~\citep{ernie5,seedream}. 
This creates a practical challenge for web-scale visual data ecosystems: AI-generated images may enter training corpora, distort real-data statistics, and introduce generation artifacts into future training cycles~\citep{longcatimage,hunyuanimage}. 
Reliable AIGC image detection is therefore becoming an important component of trustworthy visual learning systems~\citep{aigc_survey}.

The core difficulty is open-set generator shift. 
During training, a detector only observes images from a limited set of known generators. 
After deployment, it must detect images produced by newer and often stronger generators whose artifacts may differ substantially from those seen during training. 
As generators evolve, previously learned artifacts may disappear, become weaker, or be replaced by new patterns, causing the test distribution to shift over time.

Existing approaches mainly improve detector generalization through designing stronger representations or constructing better training data. Recent approaches exploit pretrained visual features~\citep{c2pclip}, frequency cues~\citep{aide}, patch-level artifacts~\citep{patchcraft}, reconstruction discrepancies~\citep{lare}, or aligned real-generated pairs~\citep{fakeinversion}. 
Although robust to certain distribution shifts, they typically follow a static train-once-and-deploy paradigm and cannot adapt to emerging generator distributions without labeled target data or retraining.

\begin{figure*}[t]
\centering
\includegraphics[width=1.0\linewidth]{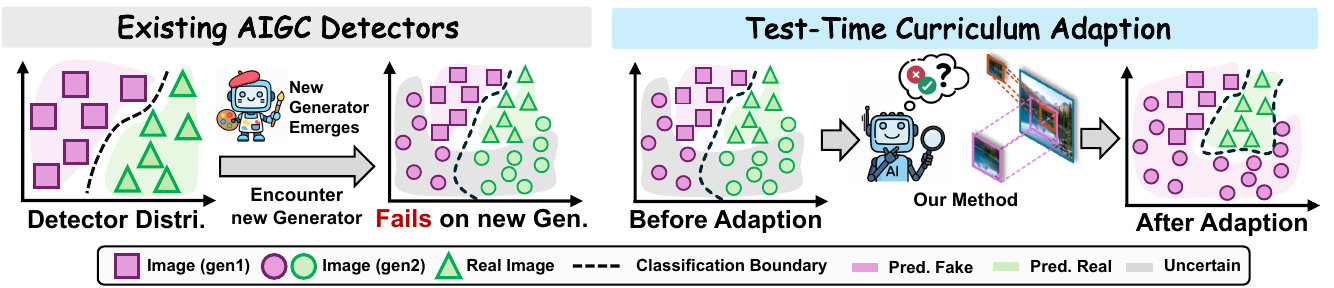}
\caption{Illustration of open-set AIGC image detection under evolving generator distributions. Existing detectors are trained on a limited set of known generators and often fail when new generators emerge, causing a mismatch between the training and test distributions. In contrast, the proposed Test-Time Curriculum (TTC) adapts the detector using unlabeled test samples, progressively shifting the decision boundary to better separate real images from images generated by unseen models.}
\label{fig:teaser}
\end{figure*}

We introduce test-time adaptation (TTA)~\citep{tta_survey,robust_tta} as a generalization paradigm for open-set AIGC detection, shifting the focus from static cross-generator generalization to adaptation on unlabeled deployment data. This paradigm allows a deployed detector to adjust to newly encountered generator distributions without collecting additional labels or retraining from scratch.
TTA is particularly attractive for industrial AIGC detection, because data curation pipelines periodically process unlabeled image collections that reflect recent generator distributions. Leveraging these data at test time can reduce the need for costly cycles of generator-specific data collection, annotation, and retraining. 

In this work, we explore a pseudo-label-based self-training~\citep{selftraining,semiself} instantiation of TTA for AIGC detection, where unlabeled test images are assigned model predictions and then used to update the detector.
However, detectors facing unseen generators may produce systematically biased predictions. Unreliable pseudo labels can introduce noise and damage the detector's original discriminative ability. 
Our key insight is that test-time adaptation should proceed as a curriculum~\citep{curriculum_survey} instead of an all-at-once process. Thus, we propose the Test-Time Curriculum (TTC), a simple and model-agnostic adaptation framework that adapts detectors through a curriculum over test samples. Our method starts from highly reliable samples and progressively expands to more informative samples. This allows the detector to adapt to the current test distribution while explicitly reducing early pseudo-label error propagation. To enable more reliable adaptation, we further integrate TTC with Cross-Scale Pseudo-Label Refinement. By aggregating complementary predictions across resolutions, this strategy reduces unreliable single-scale decisions and provides more robust supervision for curriculum adaptation. 

Finally, we observe that many existing benchmarks do not fully cover the latest high-fidelity text-to-image generators. To support evaluation under more realistic generator shifts, we construct \textbf{AIGCGuard}, a new benchmark containing 3,100 representative real images and generated counterparts from 40 recent open-source and proprietary text-to-image models, resulting in 127,100 images in total, covering diverse semantic categories, resolutions, and generator families.

In all, our contribution can be summarized as follows:
\begin{packed_lefty_item}
\item We formulate open-set AIGC image detection as a test-time adaptation problem, enabling existing detectors to adapt to unlabeled data from unseen generators without additional annotation or retraining from scratch.
\item We propose Test-Time Curriculum (TTC), a multi-round self-training framework that combines class-balanced reliable selection, hard-within-reliable sampling, parameter anchoring, and cross-scale consensus without modifying the detector architecture.
\item Comprehensive experiments demonstrate that TTC achieves consistent and significant improvements across multiple benchmarks. We contribute AIGCGuard, a modern benchmark covering 40 recent text-to-image models.
\end{packed_lefty_item}

\section{Related Work}
\subsection{AI-Generated Image Detection}
Existing AIGC image detection methods mainly follow several routes.
Early forensic methods exploit low-level spatial or frequency artifacts, such as color statistics, co-occurrence patterns, upsampling traces, and high-frequency fingerprints~\citep{aigccolor,aigc_co,fredect,npr}. 
Later works improve generalization through stronger pretrained representations~\citep{aigcdetect,univfd,fatformer,c2pclip}, patch-level analysis~\citep{patchcraft,ssp}, reconstruction discrepancy~\citep{dire, aeroblade}, and frequency modeling~\citep{safe,aide}. 
Recent dataset alignment methods further reduce training or evaluation bias by matching real and generated images in content, style, resolution, or format~\citep{aigc_bias,semgir,bfree}.  
These methods have improved cross-generator generalization, but they are usually deployed as static detectors with fixed parameters. As a result, they cannot adapt when the test distribution shifts toward unseen or stronger generators. 
In contrast to prior static detectors, we formulate open-set AIGC detection as a test-time adaptation problem and study how unlabeled target data can be used for offline-TTA.

\subsection{Test-Time Adaptation}
Test-time adaptation (TTA) adapts a pre-trained model to the target distribution during inference using only unlabeled test data. 
Existing methods mainly adapt normalization statistics~\citep{tbn,bnadapt,delta}, update model parameters with unsupervised objectives such as entropy minimization~\citep{tent,cotta}, select reliable target samples for safer updates~\citep{eata,sar}, or construct target-domain prototypes and caches~\citep{t3a,tda}. 
However, TTA remains underexplored for AIGC image detection, where unseen generator shifts can produce biased pseudo labels and unstable adaptation. We address this gap with a curriculum-based adaptation strategy that starts from balanced reliable pseudo-labeled samples and progressively incorporates harder yet informative cases.
\section{Method}

\begin{figure*}[t]

\centering

\includegraphics[width=0.9\linewidth]{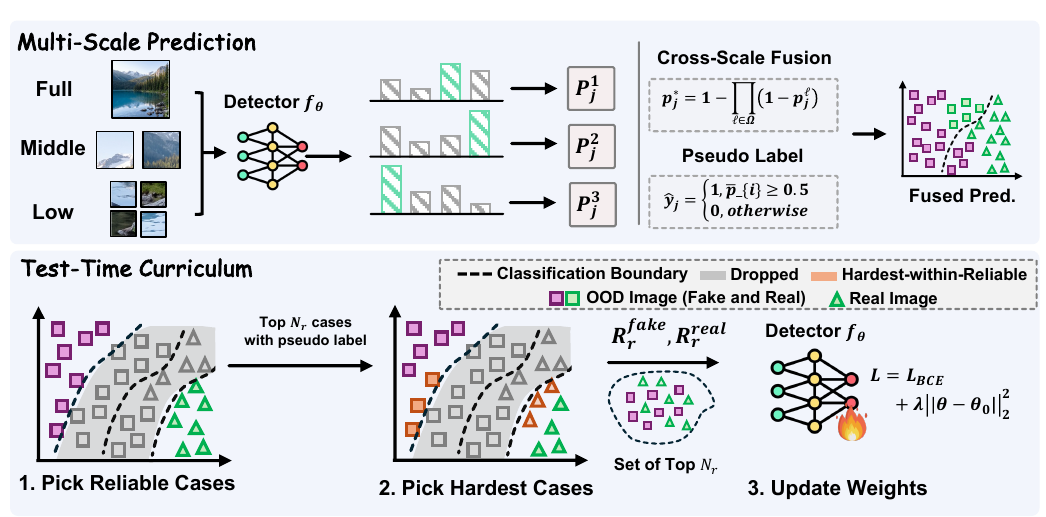}

\caption{Overview of the proposed TTC framework. Given unlabeled test images, TTC first performs multi-scale prediction and cross-scale fusion to obtain more reliable pseudo labels. It then applies a curriculum-based adaptation procedure with two stages: selecting reliable pseudo-labeled samples and further prioritizing hard-within-reliable examples for informative updates. The detector is updated iteratively across rounds, while noisy-or fusion is used at inference to aggregate evidence across scales for final prediction.}

\label{fig:method}

\end{figure*}

We first present \textbf{(1) Test-Time Curriculum Adaptation Framework (TTC)} as the overall framework for adapting detectors at test time. Then, we introduce a \textbf{(2) Cross-Scale Pseudo-Label Refinement} to enhance TTC. Together, these two components form a practical pipeline for robust adaptation under evolving AIGC test distributions. The overall framework is shown in Fig.~\ref{fig:method}.

\subsection{Preliminary}
The task is to distinguish AI-generated images from real images in an open-set deployment scenario. During training, a labeled set \(\mathcal{D}_{train} = \{(x_i, y_i)\}_{i=1}^{N}\) is drawn from the training distribution \(p_{\text{train}}(x)\), where \(N\) denotes the number of training samples and \(y_i \in \{0,1\}\) denotes the real/fake label. Fake training images are generated by a limited set of known generators \(\{G_1, \dots, G_K\}\). At test time, the detector is given an unlabeled test set \(\mathcal{D}_{test} = \{x_j\}_{j=1}^{M}\), where \(M\) denotes the number of test samples and \(x_j \sim p_{\text{test}}(x)\). The test set may contain images from previously unseen generators \(\{G_{K+1}, \dots\}\), leading to a distribution shift \(p_{\text{test}}(x) \neq p_{\text{train}}(x)\). As new generators emerge, \(p_{\text{test}}(x)\) may further evolve during deployment.

The objective is to learn a detector \(f_\theta: x \mapsto y \in \{0,1\}\) that generalizes to unseen generators under the shifted test distribution \(p_{\text{test}}(x)\). In a test-time adaptation (TTA) setting~\citep{eata,cycleself}, the initial pretrained detector \(f_\theta\) is allowed to update its parameters using the unlabeled test set \(\mathcal{D}_{test}\), without access to ground-truth labels. In this work, TTA is instantiated as pseudo-label-based self-training. At the beginning of adaptation, the pretrained detector \(f_\theta\) generates pseudo labels \(\hat{y}_j\) for unlabeled test samples \(x_j\). A generic self-training update at the adaptation stage \(t\) can be written as
\begin{equation}
\theta' = \theta - \eta \nabla_\theta \mathcal{L}(f_{\theta}(x_j), \hat{y}_j),
\quad x_j \in \mathcal{D}_{test},
\end{equation}
where \(\mathcal{L}(\cdot)\) is a loss function, and \(\eta\) is the learning rate. 
Due to distribution shifts, the detector \(f_\theta\) may produce incorrect predictions, which manifest as noise in the pseudo labels and can degrade model performance. To mitigate this issue, we propose Test-Time Curriculum (TTC), which explicitly controls error propagation during adaptation.

\subsection{Test-Time Curriculum Adaptation Framework}

\noindent \textbf{Pseudo-Labeling Selection.}
We first focus on selecting reliable pseudo-labeled samples at curriculum round \(r\), which are critical for safe adaptation under distribution shift. A reliable sample is one for which the detector's prediction is likely correct, thereby minimizing the risk of propagating early errors under distribution shift. To quantify reliability, we introduce the binary entropy \(H(\cdot)\)~\citep{bientropy} of the predicted probability \(p_j\) for each test sample:
\begin{equation}
H(p_j) = -p_j \log p_j - (1-p_j) \log (1-p_j),
\end{equation}
where \(p_j = f_{\theta}(x_j)\) denotes the predicted probability of sample \(x_j\) being generated. Lower entropy indicates a prediction farther from the decision boundary and thus more trustworthy. Based on this reliability measure, TTC selects samples whose entropy falls below an uncertainty threshold \(\tau_r\):
\begin{equation}
\mathcal{R}_r = \{x_j \in \mathcal{D}_{test} \mid H(p_j) \le \tau_r \}.
\end{equation}
This set contains both reliably predicted real and fake samples. Within \(\mathcal{R}_r\), samples are separated into two classes based on the predicted label: the reliable fake candidates \(\mathcal{R}_r^{\text{fake}}\) and the reliable real candidates \(\mathcal{R}_r^{\text{real}}\). Formally,
\begin{equation}
\begin{aligned}
\mathcal{R}_r^{\text{fake}}
&= \left\{x_j \in \mathcal{R}_r \mid \hat{y}_j = 1\right\}, \\
\mathcal{R}_r^{\text{real}}
&= \left\{x_j \in \mathcal{R}_r \mid \hat{y}_j = 0\right\}, \\
\hat{y}_j
&= \mathbf{1}\!\left[p_j > 0.5\right].
\end{aligned}
\end{equation}
In practice, real images are often easier to classify confidently, while images from newly emerging generators tend to produce higher uncertainty. 
Because entropy-based selection relies on prediction uncertainty rather than the number of samples, the numbers of reliably predicted real and fake samples may therefore differ, \ie, \(|\mathcal{R}_r^{\text{real}}| \neq |\mathcal{R}_r^{\text{fake}}|\), especially in early curriculum rounds with strict thresholds. 
Such imbalance can induce class-wise gradient bias, causing the adaptation to be dominated by the over-represented class and potentially degrading performance. 
To enforce class balance, TTC selects an equal number of samples from each class. 
Specifically, the number of samples per class is given by \(N_r = \min(|\mathcal{R}_r^{\text{fake}}|, |\mathcal{R}_r^{\text{real}}|)\), ensuring that both classes contribute equally to the adaptation. 
Class balancing is used as an optimization constraint rather than as an estimate of the target class prior. The original target distribution is not reweighted at evaluation time; balancing only prevents the adaptation gradient from being dominated by one predicted class.
If either reliable class is empty, the current round is skipped and the detector retains the last safe parameters, preventing degenerate updates on effectively single-class target batches. 

Once class balance is established, a new challenge arises: which samples within each equally sized class subset should be used for adaptation to maximize learning? To this end, TTC performs a second-stage selection within each class, focusing on the informative samples. 
Although low-uncertainty samples in \(\mathcal{R}_r\) are reliable, they are often already well-classified and provide limited new information for adaptation. 
To focus adaptation on informative samples, TTC selects a hard-within-reliable subset \(\mathcal{S}_r\) from the reliable set \(\mathcal{R}_r\), prioritizing samples with higher uncertainty within each class:
\begin{equation}
\mathcal{S}_r = 
\underbrace{\operatorname*{arg\,max}_{\substack{S \subseteq \mathcal{R}_r^{\text{fake}} \\ |S| = N_r}} \sum_{x_j \in S} H(p_j)}_{\text{hard fake samples}}
\;\cup\;
\underbrace{\operatorname*{arg\,max}_{\substack{S \subseteq \mathcal{R}_r^{\text{real}} \\ |S| = N_r}} \sum_{x_j \in S} H(p_j)}_{\text{hard real samples}},
\end{equation}
where samples closest to the decision boundary are prioritized within each class. This ensures adaptation focuses on challenging yet trustworthy examples, improving the detector's ability to capture subtle cues in generated images.

The above selective sampling rule separates label reliability from learning difficulty: the two confidence regions provide reliable pseudo labels, hard-within-reliable sampling focuses adaptation on informative samples. Balanced selection controls class-wise gradient bias. High-entropy samples with predictions near 0.5 are excluded from the reliable set, reducing the risk of propagating early errors. 
This set \(\mathcal{S}_r\) then forms the input for the curriculum update in round \(r\), allowing the detector to progressively incorporate harder samples as it adapts to the evolving test distribution \(p_{\text{test}}\).

\noindent \textbf{Curriculum Adaptation.}
In the presence of distribution shifts, directly adapting a detector to a new, unseen test distribution \(p_{\text{test}}\) can be challenging due to the limited availability of reliable samples in the early stages of adaptation. 
To address this, TTC adopts a curriculum approach, where the detector is gradually adapted in multiple rounds using progressively relaxed uncertainty thresholds. The first round focuses on the most reliable samples with a strict threshold \(\tau_0\), as the model is still distant from the test distribution and the available trustworthy data \(\mathcal{S}_0\) is limited. In subsequent rounds, the threshold is relaxed, allowing the detector to incorporate harder, more uncertain samples that are closer to the decision boundary. As each round brings the model closer to \(p_{\text{test}}\), the set of reliable samples expands, enabling more comprehensive adaptation.

For the selected hard-within-reliable set \(\mathcal{S}_r\) in round \(r\), TTC updates the detector using pseudo-label supervision~\citep{bce} while constraining deviation from the initial pretrained parameters \(\theta_0\). 
The loss function for round \(r\) is
\begin{equation}
\mathcal{L}^{(r)} = \text{BCE}(f_\theta(x_j), \hat{y}_j^{(r)}) + \lambda \|\theta - \theta_0\|_2^2, \quad x_j \in \mathcal{S}_r,
\end{equation}
where \(\lambda\) controls the strength of L2 regularization~\citep{l2}, preventing catastrophic drift from noisy pseudo labels while enabling stable adaptation.

After each round, the detector recomputes the predicted probabilities for all test samples, updating the selected subset for the next round. 
This iterative process forms a curriculum: reliable samples guide the initial adaptation, and the adapted detector progressively incorporates harder, moderately uncertain samples. This progressive design is intended to reduce the influence of uncertain pseudo labels in early rounds while gradually expanding the adaptation set as the detector becomes better aligned with the target distribution, while a label-free safety check rejects unstable rounds and restores the last safe model when necessary.

\begin{table*}[t]
\centering
\begin{adjustbox}{max width=\textwidth}
\begin{tabular}{lcccccccccccccccccl}
\toprule
Method & \rotatebox{90}{ADM} & \rotatebox{90}{DALLE2} & \rotatebox{90}{GLIDE} & \rotatebox{90}{Midjourney} & \rotatebox{90}{VQDM} & \rotatebox{90}{BigGAN} & \rotatebox{90}{CycleGAN} & \rotatebox{90}{GauGAN} & \rotatebox{90}{ProGAN} & \rotatebox{90}{SDXL} & \rotatebox{90}{SD14} & \rotatebox{90}{SD15} & \rotatebox{90}{StarGAN} & \rotatebox{90}{StyleGAN} & \rotatebox{90}{StyleGAN2} & \rotatebox{90}{WFR} & \rotatebox{90}{Wukong} & \rotatebox{90}{Avg.} \\
\midrule
UnivFD & 62.5 & 50.0 & 61.3 & 55.1 & 76.9 & 87.5 & \underline{96.9} & 98.8 & \textbf{99.4} & 58.2 & 55.6 & 55.7 & 95.1 & 80.0 & 69.4 & 69.2 & 61.1 & 72.5 \\
NPR & 43.8 & 20.0 & 41.2 & 53.4 & 48.4 & 53.1 & 76.6 & 42.2 & 58.7 & 59.6 & 55.1 & 55.0 & 67.4 & 57.9 & 54.6 & 58.8 & 57.4 & 53.1 \\
FatFormer & 80.2 & 68.5 & \underline{91.1} & 54.4 & 88.0 & \textbf{99.2} & \textbf{99.5} & \textbf{99.1} & 98.5 & 71.7 & 67.5 & 67.2 & \underline{99.4} & \textbf{98.0} & \textbf{98.8} & 88.3 & 75.6 & 85.0 \\
SAFE & 49.5 & 49.5 & 53.0 & 49.0 & 50.2 & 52.2 & 51.9 & 50.0 & 50.0 & 49.8 & 49.7 & 49.8 & 50.1 & 50.0 & 50.0 & 49.8 & 50.3 & 50.3 \\
DRCT & 79.9 & 89.2 & 89.2 & 85.5 & \underline{88.6} & 81.4 & 91.0 & 93.8 & 71.1 & 88.3 & 91.4 & 91.0 & 53.0 & 62.7 & 63.8 & 73.9 & 90.8 & 81.4 \\
AIDE & 52.9 & 51.1 & 60.2 & 49.8 & 69.3 & 70.1 & 93.6 & 60.6 & 89.0 & 49.6 & 51.6 & 51.0 & 72.1 & 66.5 & 59.0 & 80.6 & 54.5 & 63.6 \\
C2P-CLIP & 71.6 & 52.3 & 73.5 & 56.6 & 73.7 & \underline{98.4} & 96.8 & 98.8 & \underline{99.3} & 62.3 & 77.5 & 76.9 & \textbf{99.6} & \underline{93.1} & 79.4 & \textbf{94.8} & 79.4 & 81.4 \\
AlignedForensics & 51.6 & 52.0 & 55.6 & \underline{96.2} & 72.1 & 51.2 & 49.5 & 50.8 & 50.7 & 95.1 & \textbf{99.7} & \textbf{99.6} & 53.8 & 52.7 & 51.6 & 50.0 & \textbf{99.6} & 66.6 \\
DDA & \underline{89.5} & \underline{94.6} & 89.6 & 95.6 & 76.6 & 91.0 & 72.5 & 92.7 & 92.8 & \textbf{99.4} & 98.7 & 98.6 & 72.7 & 87.8 & \underline{90.2} & 52.1 & 98.8 & \underline{87.8} \\
\midrule
Base & 70.1 & 88.5 & 73.2 & 66.5 & 71.5 & 90.7 & 60.0 & \underline{99.0} & 87.4 & 97.2 & \underline{99.1} & \underline{99.0} & 49.4 & 80.2 & 75.9 & 81.0 & \underline{99.1} & 81.6 \\
\rowcolor{tableblue} \textbf{Ours} & \textbf{89.7} & \textbf{99.2} & \textbf{95.7} & \textbf{98.7} & \textbf{98.4} & 93.7 & 69.4 & 97.4 & 97.2 & \underline{99.3} & 98.7 & 98.8 & 76.6 & 90.9 & 87.1 & \underline{91.9} & 98.8 & \textbf{93.0} \\
\bottomrule
\end{tabular}
\end{adjustbox}
\caption{Comparison on the AIGCDetect Benchmark. We report balanced accuracy (\%) for each generator subset and the average across subsets. ``Base'' denotes the detector initialized with the same backbone and training data as DDA for fair comparison, but without using DDA's alignment strategy. ``Ours'' applies the proposed TTC adaptation on top of this Base detector. The best result and the second-best result are marked in \textbf{bold} and \underline{underline}, respectively.}
\label{tab:aigcdetection}
\end{table*}

\subsection{Cross-Scale Pseudo-Label Refinement}
Effective pseudo-labeling for Test-Time Curriculum (TTC) requires capturing artifacts present at different visual granularities, from fine-grained local traces~\citep{allpatch} to coarse structural patterns~\citep{c2pclip}. Single-scale predictions often fail to detect all relevant cues, leading to noisy supervision. We therefore introduce a cross-scale refinement strategy that combines multi-resolution predictions, improving pseudo-label reliability for curriculum-based adaptation.

\noindent \textbf{Multi-Scale Prediction.}
Artifacts introduced by AI-generated images often appear at different visual granularities, ranging from subtle local traces (e.g., high-frequency noise, edges) to coarse structural or semantic inconsistencies. 
To capture such multi-scale evidence, each test image \(x_j\) is processed at a predefined set of scales \(\Omega\), which may include the full image, intermediate downsampled resolutions, or localized patches extracted from regions likely to contain artifacts. 

For patch-based scales, informative patches \(\{x_{j,k}^\ell\}_{k=1}^{N_\ell}\) are selected based on simple heuristics such as local gradient magnitude, high-frequency energy, and texture complexity, while also including a small number of random patches to preserve spatial diversity.

At each scale \(\ell \in \Omega\), the detector produces a patch-level probability \(p_{j,k}^\ell\) for each patch. These patch-level probabilities are aggregated into a single scale-level probability \(p_j^\ell\) using logit-space log-sum-exp pooling~\citep{lse,mil_survey}:
\begin{equation}
p_j^\ell = \sigma(T_\ell \log \frac{1}{N_\ell} \sum_{k=1}^{N_\ell} \exp\left(\frac{\sigma^{-1}(p_{j,k}^\ell)}{T_\ell}\right)).
\end{equation}
Here, \(\sigma^{-1}(\cdot)\) denotes the inverse sigmoid function (logit), \(\sigma(\cdot)\) is the sigmoid function mapping logit back to probability. \(T_\ell\) is a hyperparameter controlling the selectivity of aggregation. This aggregation preserves strong localized evidence while allowing multiple moderately suspicious patches to contribute, improving the reliability of individual scale predictions.

Importantly, this multi-scale prediction framework is model-agnostic: it can be applied to any pre-trained detector without modifying its architecture or training procedure. By extracting complementary evidence across multiple resolutions~\citep{multiscale,crossscale}, it provides a robust foundation for the subsequent Cross-Scale Pseudo-Label Refinement, improving pseudo-label reliability for Test-Time Curriculum (TTC) adaptation.

\noindent \textbf{Cross-Scale Pseudo-Label Refinement.}
Single-scale predictions, even when aggregated at the patch level, may still be unreliable due to calibration inconsistencies across scales or variations in generator artifacts. 
To improve the reliability of pseudo labels for Test-Time Curriculum adaptation, we aggregate predictions across multiple independent scales to form a cross-scale mean probability \(\bar{p}_j = \frac{1}{|\Omega|}\sum_{\ell \in \Omega} p_j^{\ell},\) where \(p_j^{\ell}\) are the scale-level probabilities computed from multi-scale predictions. This fused probability is used for selecting reliable pseudo labels~\citep{mean}, replacing single-scale probabilities when forming candidate sets for real and fake samples.

At curriculum round \(r\), the cross-scale mean \(\bar{p}_j\) is used both to compute prediction entropy and to assign a hard pseudo label, \(\hat{y}_j = \mathbf{1}\!\left[p_j > 0.5\right]\). The BCE loss is optimized using this hard label. 

For final inference, we combine all scales using noisy-or fusion~\citep{noisyor}:
\begin{equation}
s_j
= 1-\prod_{\ell \in\Omega}(1-p_j^{\ell}).
\end{equation}
Because predictions from different scales are correlated, \(s_j\) is not interpreted as a calibrated posterior probability. We use it only as a fixed heuristic aggregation score that emphasizes images receiving a strong fake response at any scale. In contrast, pseudo-label reliability and pseudo-label assignment are determined using the cross-scale mean 
\(\bar{p}_j\), which is less sensitive to an isolated high-response scale.


Finally, in each TTC round, the cross-scale mean probability \(\bar{p}_j\) is used as the fused pseudo-label target, while scale-specific probabilities guide the selection of informative samples for adaptation.
The shared cross-scale target encourages predictions obtained from different resolutions to agree with their aggregate consensus, maintaining stable performance across self-training rounds.
\section{Experiments}


\begin{table*}[t]
\centering
\begin{adjustbox}{max width=\textwidth}
\begin{tabular}{lccccccccccl}
\toprule
Method & Janus & J-Pro-1B & J-Pro-7B & Show-o & LlamaGen & Infinity & VAR & PixArt-XL & SD3.5-L & FLUX & Avg. \\
\midrule
UnivFD & 70.6 & 68.3 & 59.4 & 57.9 & 75.2 & 56.7 & 66.7 & 53.9 & 74.0 & 51.6 & 63.4 \\
NPR & 30.1 & 29.8 & 29.8 & 79.6 & 79.5 & 79.8 & 79.3 & 78.7 & 79.4 & 79.1 & 64.5 \\
FatFormer & 48.8 & 48.6 & 48.8 & 73.2 & 84.3 & 53.9 & 90.1 & 67.1 & 71.0 & 54.6 & 64.0 \\
SAFE & 49.9 & 50.0 & 50.0 & \textbf{99.7} & \underline{99.4} & \textbf{99.9} & 95.3 & \textbf{99.7} & 95.7 & \textbf{99.4} & 83.9\\
DRCT & 71.1 & 71.7 & 70.9 & 70.0 & 70.8 & 71.3 & 61.2 & 70.8 & 69.6 & 67.2 & 69.5 \\
AIDE & 50.6 & 52.6 & 51.5 & 98.7 & \textbf{99.7} & \underline{99.7} & \underline{98.6} & 98.8 & \underline{99.0} & 95.4 & 84.5 \\
C2P-CLIP & 72.4 & 86.9 & 86.0 & 99.0 & 98.8 & 99.1 & 94.0 & 98.9 & 98.6 & 96.1 & 93.0 \\
AlignedForensics & 93.9 & 92.3 & 92.4 & 81.3 & 94.0 & 94.9 & 89.3 & 95.1 & 84.9 & 92.0 & 91.0  \\
DDA & \underline{98.5} & \textbf{99.3} & \underline{98.7} & 94.8 & 99.3 & 98.9 & 80.4 & \underline{99.4} & 97.0 & 97.2 & \underline{96.3} \\
\midrule
Base & 95.5 & \underline{98.8} & 97.3 & 95.4 & 98.0 & 64.4 & 84.1 & 94.6 & 50.5 &  50.6 & 82.9 \\
\rowcolor{tableblue} \textbf{Ours} & \textbf{99.2} & \textbf{99.3} & \textbf{99.1} & \underline{99.1} & 99.2 & 99.2 & \textbf{99.0} & 99.0 & \textbf{99.2} & \underline{99.1} & \textbf{99.1} \\
\bottomrule
\end{tabular}
\end{adjustbox}
\caption{Comparison on the AIGI-Holmes Benchmark. The best result and the second-best result are marked in \textbf{bold} and \underline{underline}, respectively.}
\label{tab:aigi}
\end{table*}

\subsection{Experimental Setup}
\noindent \textbf{Dataset.} 
We evaluate on five evaluation suites: three established benchmark collections (AIGCDetectionBenchmark~\citep{aigcdetect}, AIGI-Holmes~\citep{aigihlomes} and GenImage~\citep{genimage}), one in-the-wild collection, Chameleon~\citep{aide}, and our controlled modern-generator benchmark, AIGCGuard. These datasets comprise real images from multiple sources and fake images generated by a variety of generative models, including diffusion models, GANs, auto-regressive models, and other unseen generators. The datasets differ in format, content, resolution and curation procedures, providing a comprehensive evaluation. Detailed curation procedures for AIGCGuard are provided in the Appendix.

Unless otherwise stated, we evaluate mixed-domain offline batch TTA. For each benchmark, TTC receives the full unlabeled target split as a single mixed-generator batch and performs one adaptation process without access to generator identities or target labels. The adapted detector is then evaluated on the same target collection, making the primary protocol transductive. To examine whether improvements are caused mainly by fitting the adaptation images, Table 10 additionally adapts on a sampled subset and evaluates on the non-overlapping remainder after removing all adaptation samples. The difference between full-set and non-overlapping evaluation is at most 0.10 BAcc in this analysis. Generator-specific adaptation is reported only as a diagnostic protocol in the Appendix.

\noindent \textbf{Implementation Details.}
TTC is model-agnostic and can be applied to different AIGC detectors. For fair comparison, we use the same pretrained backbone and training data as DDA, but do not apply its data-alignment strategy; this detector is used as the Base model for TTC adaptation. Unless otherwise stated, TTC is adapted once on the full unlabeled test split of each benchmark without generator identities. We use 3 curriculum rounds with confidence thresholds (0.9, 0.8, 0.7), converted to corresponding binary-entropy thresholds \(\tau\) for reliable pseudo-label selection. The detector is updated using SGD with a learning rate of 5e-4 and momentum of 0.9, and 3 adaptation epochs. All benchmarks use the same hyperparameters without dataset-specific tuning. Following reliable and shift-aware TTA practices~\citep{ttasafe,ttn,ttalabel}, we apply a label-free safety check to skip or roll back adaptation when prediction statistics indicate severe drift or confidence collapse. Multi-scale prediction uses three input scales: \(70 \times 70\), \(224 \times 224\), and \(336 \times 336\) with 24, 6, and 1 patch per image, respectively. The LSE pooling temperature is set to \(T_\ell=1.5\). More details are provided in the Appendix.

\noindent \textbf{Evaluation Metrics.}
Following common practice in AIGC detection, we report balanced accuracy—the average of real and fake image accuracies—as the evaluation metric to ensure fair comparison across potentially skewed datasets. This metric computes the unweighted arithmetic mean of the individual detection accuracies for authentic and generated images, providing a comprehensive and unbiased measure of each model’s fundamental discriminative capability.

\noindent \textbf{Comparative Methods.}
We evaluate 9 advanced AIGC detection methods covering a range of representative approaches. The comparative baselines include data-alignment-based methods such as DDA~\citep{dda}, DRCT~\citep{drct}, and AlignedForensics~\citep{alignedforensics}; advanced frequency-based detectors including NPR~\citep{npr}, SAFE~\citep{safe}, and AIDE~\citep{aide}; and vision-language model-based approaches such as UnivFD~\citep{univfd}, Fatformer~\citep{fatformer}, and C2P-CLIP~\citep{c2pclip}. 
To isolate the contribution of the proposed adaptation strategy, we further compare TTC with representative general-purpose TTA methods. Additional analyses of detector generalization, adaptation efficiency, disjoint adaptation--evaluation splits, hyperparameter sensitivity, and run-to-run stability are provided in the Appendix.

\begin{figure*}[!t]
\centering
\includegraphics[width=0.9\textwidth]{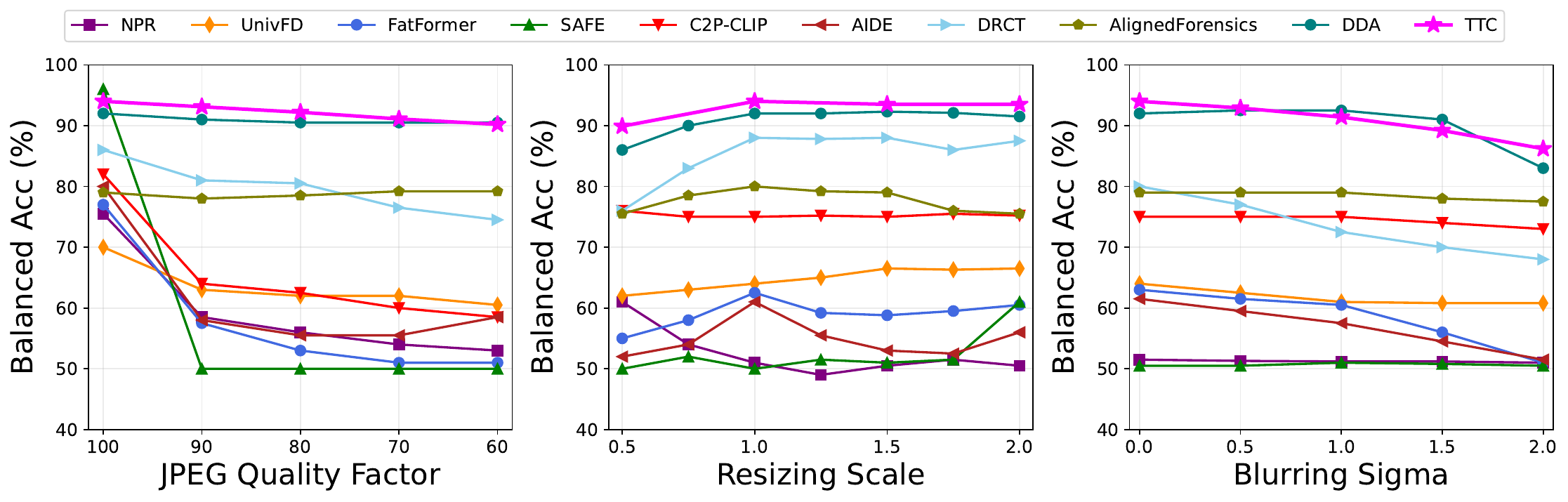}
\caption{Robustness under varying JPEG compression, resizing, and Gaussian blur on GenImage. The experiment evaluates adaptation to changes in perturbation severity and mixture.}
\label{fig:robustness}
\end{figure*}

\subsection{Main Results}

\noindent \textbf{Comparison on AIGCDetection.}

As shown in Table~\ref{tab:aigcdetection}, TTC achieves the best average balanced accuracy of 93.0\%, outperforming the Base model by 11.4 points and DDA by 5.2 points. TTC delivers especially large gains on challenging generators such as Midjourney (+32.2), StarGAN (+27.2), VQDM (+26.9), GLIDE (+22.5), and ADM (+19.6) over the Base model. It also surpasses DDA substantially on several difficult cases, including WFR (+39.8) and VQDM (+21.8). These results indicate that TTC effectively adapts the detector to generator shifts beyond the fixed training distribution.

\noindent \textbf{Comparison on AIGI-Holmes.}

Table~\ref{tab:aigi} shows that TTC achieves the best average balanced accuracy of 99.1\%, improving over the Base model by 16.2 points and over DDA by 2.8 points. The gains are particularly notable on challenging subsets where the Base detector degrades severely, including Infinity (64.4\%\(\rightarrow\)99.6\%), SD3.5-L (50.5\%\(\rightarrow\)99.2\%), and FLUX (50.6\%\(\rightarrow\)99.1\%). TTC also achieves near-saturated performance across diverse generator families, including autoregressive, diffusion, and multimodal generators, with all subset accuracies reaching at least 99.0\%. These results show that TTC remains effective under mixed-generator adaptation without relying on generator identities.

\begin{table}[t]
\centering
      \begin{adjustbox}{max width=\linewidth}
      \begin{tabular}{lcc}
      \toprule
      & BAcc (\%) & \(\Delta\) \\
      \midrule
Base & 82.9 & -- \\
\midrule
+ BRS & 92.9 & +10.0 \\
+ HWR & 83.8 & +0.9 \\
+ CSPLR & 97.0 & +14.1 \\
\midrule
Full & 99.1 & +15.6 \\
\bottomrule
      \end{tabular}
      \end{adjustbox}
    \caption{Ablation study on the AIGI-Holmes Benchmark. BRS denotes Balanced Reliable Selection, and HWR denotes Hard-within-Reliable Selection, and CSPLR denotes Cross-Scale Pseudo-Label Refinement. Rows are cumulative: “+HWR” includes BRS, “+CSPLR” includes both BRS and HWR, and “Full” additionally applies noisy-or score fusion at inference.}
    \label{tab:ablation}
\end{table}

\noindent \textbf{Comparison on Chameleon and AIGCGuard.}

We further evaluate TTC on Chameleon, an in-the-wild collection, and AIGCGuard, a controlled prompt-matched benchmark covering recent high-fidelity generators.
As shown in Table~\ref{tab:in-the-wild}, TTC improves over the strongest previous baseline DDA from 82.4\% to 91.1\% on Chameleon. On the more challenging AIGCGuard benchmark, where many existing detectors suffer substantial performance drops, TTC achieves the best result of 78.4\%, outperforming both the Base model and all compared baselines. These results highlight the advantage of adapting to evolving real-world test distributions at test time.

\subsection{Ablation Studies}
We conduct ablation studies on the AIGI-Holmes benchmark to evaluate the contribution of each component in TTC. 
Table~\ref{tab:ablation} reveals a strong interaction between sample difficulty and pseudo-label reliability. Balanced reliable selection (BRS) provides most of the gain in the single-scale setting, improving BAcc from 82.9\% to 92.9\%. Adding HWR without cross-scale refinement reduces performance to 83.8\%, because HWR deliberately emphasizes samples closer to the decision boundary and is therefore more sensitive to pseudo-label noise. Once CSPLR is introduced, the same hard-within-reliable strategy reaches 97.0\%, indicating that informative sample selection becomes effective when pseudo-label reliability is sufficiently controlled. Noisy-or contributes a further inference-time improvement to 99.1\%. We therefore interpret Table~\ref{tab:ablation} as evidence of component complementarity rather than attributing the full improvement to HWR or the curriculum schedule alone.

\subsection{Robustness to Perturbations}

We evaluate robustness on GenImage under common image perturbations, including JPEG compression, resizing, and Gaussian blur. 
This experiment does not evaluate robustness to entirely unseen corruption families. Instead, it examines whether target-time adaptation can accommodate changes in the mixture and severity of these perturbations. 

As shown in Fig.~\ref{fig:robustness}, TTC maintains the strongest performance across most evaluated severity levels, indicating that adaptation remains effective when the target batch contains heterogeneous image degradation.

\begin{table}[t]
\centering
    \begin{adjustbox}{max width=\linewidth}
    \begin{tabular}{lcc}
    \toprule
Method & Chameleon & AIGCGuard \\
\cmidrule(lr){1-1} \cmidrule(l){2-3}
UnivFD & 50.7 & 50.4 \\
NPR & 59.9 & 56.6 \\
FatFormer & 51.2 & 50.6 \\
SAFE & 59.2 & 49.7 \\
DRCT & 56.6 & 47.5 \\
AIDE & 63.1 & 54.4 \\
C2P-CLIP & 51.1 & 55.5 \\
AlignedForensics & 71.0 & 60.8 \\
DDA & \underline{82.4} & \textbf{77.2} \\
\midrule
Base & 62.4 & 60.6 \\
\rowcolor{tableblue} \textbf{Ours} & \textbf{91.1} & \textbf{78.4} \\
\bottomrule
    \end{tabular}
    \end{adjustbox}
        \caption{Comparison on the Chameleon Benchmark and our AIGCGuard Benchmark.}
        \label{tab:in-the-wild}

\end{table}

\section{Conclusion}
We present TTC, a simple and model-agnostic test-time adaptation framework for open-set AIGC image detection. TTC improves adaptation stability under generator shift through balanced reliable selection, hard-within-reliable curriculum updates, and cross-scale pseudo-label refinement. We also introduce AIGCGuard, a new benchmark built from recent high-fidelity text-to-image models for evaluating realistic open-set detection. Extensive experiments show that TTC substantially improves detection performance across diverse benchmarks and unseen-generator settings. These results suggest that test-time adaptation is a practical and effective direction for maintaining AIGC detector robustness in rapidly evolving real-world environments.


\appendix
\section{Technical appendices and supplementary material}

\begin{table*}[t]
\centering
\begin{adjustbox}{max width=\textwidth}
\begin{tabular}{lccccccccl}
\toprule
Method & Midjourney & SDv1.4 & SDv1.5 & ADM & GLIDE & Wukong & VQDM & BigGAN & Avg. \\
\midrule
UnivFD & 55.1 & 55.6 & 55.7 & 62.5 & 61.3 & 61.1 & 76.9 & 84.4 & 64.1 \\ 
NPR & 53.4 & 55.1 & 55.0 & 43.8 & 41.2 & 57.4 & 48.4 & 57.7 & 51.5 \\
FatFormer & 52.1 & 53.6 & 53.8 & 61.4 & 65.5 & 60.9 & 72.5 & 82.2 & 62.8 \\
SAFE & 49.0 & 49.7 & 49.8 & 49.5 & 53.0 & 50.3 & 50.2 & 50.9 & 50.3 \\
DRCT & 82.4 & 88.3 & 88.2 & 76.9 & 86.1 & 87.9 & \underline{85.4} & \textbf{87.0} & 84.7 \\
AIDE & 58.2 & 77.2 & 77.4 & 50.4 & 54.6 & 70.5 & 50.8 & 50.6 & 61.2 \\
C2P-CLIP & 56.6 & 77.5 & 76.9 & 71.6 & 73.5 & 79.4 & 73.7 & 85.9 & 74.4 \\
AlignedForensics & \underline{97.5} & \textbf{99.7} & \textbf{99.6} & 52.4 & 57.6 & \textbf{99.6} & 75.0 & 50.6 & 79.0 \\
\textbf{DDA} & 95.6 & 98.7 & 98.6 & \textbf{89.5} & \underline{89.6} & 98.7 & 76.5 & \underline{86.5} & \underline{91.7} \\
\midrule
Base & 66.5 & \underline{99.1} & \underline{99.0} & 70.1 & 73.2 & \underline{99.1} & 71.5 & 62.0 & 80.1 \\ 
\rowcolor{tableblue} \textbf{Ours} & \textbf{98.6} & 98.9 & 98.8 &\underline{87.6} & \textbf{94.6} & 98.8 & \textbf{98.2} & 64.8 & \textbf{92.5} \\
\bottomrule
\end{tabular}
\end{adjustbox}
\caption{Comparison on the GenImage Benchmark. We report balanced accuracy (\%) for each generator subset and the average across subsets. The best result and the second-best result are marked in \textbf{bold} and \underline{underline}, respectively.}
\label{tab:genimage}
\end{table*}



\subsection{Results on GenImage}

We further evaluate TTC on the GenImage benchmark under the same mixed-domain adaptation protocol. As shown in Table~\ref{tab:genimage}, TTC achieves the best average balanced accuracy of 92.5\%, improving over the Base detector by 12.4 points and outperforming all baselines.
TTC brings particularly large gains on generators where the Base detector generalizes poorly, such as Midjourney, GLIDE, and VQDM, improving balanced accuracy from 66.5\% to 98.6\%, from 73.2\% to 94.6\%, and from 71.5\% to 98.2\%, respectively. These results show that TTC can effectively adapt to mixed unseen generator distributions and improve overall performance on GenImage using only unlabeled test data.

\begin{table*}[t]
\centering
\begin{adjustbox}{max width=\textwidth}
\begin{tabular}{lcccccccccccccccccl}
\toprule
Method & \rotatebox{90}{ADM} & \rotatebox{90}{DALLE2} & \rotatebox{90}{GLIDE} & \rotatebox{90}{Midjourney} & \rotatebox{90}{VQDM} & \rotatebox{90}{BigGAN} & \rotatebox{90}{CycleGAN} & \rotatebox{90}{GauGAN} & \rotatebox{90}{ProGAN} & \rotatebox{90}{SDXL} & \rotatebox{90}{SD14} & \rotatebox{90}{SD15} & \rotatebox{90}{StarGAN} & \rotatebox{90}{StyleGAN} & \rotatebox{90}{StyleGAN2} & \rotatebox{90}{WFR} & \rotatebox{90}{Wukong} & \rotatebox{90}{Avg.} \\
\midrule
UnivFD & 62.5 & 50.0 & 61.3 & 55.1 & 76.9 & 87.5 & \underline{96.9} & 98.8 & \textbf{99.4} & 58.2 & 55.6 & 55.7 & 95.1 & 80.0 & 69.4 & 69.2 & 61.1 & 72.5 \\
NPR & 43.8 & 20.0 & 41.2 & 53.4 & 48.4 & 53.1 & 76.6 & 42.2 & 58.7 & 59.6 & 55.1 & 55.0 & 67.4 & 57.9 & 54.6 & 58.8 & 57.4 & 53.1 \\
FatFormer & 80.2 & 68.5 & 91.1 & 54.4 & 88.0 & \textbf{99.2} & \textbf{99.5} & \underline{99.1} & 98.5 & 71.7 & 67.5 & 67.2 & \underline{99.4} & \textbf{98.0} & \textbf{98.8} & 88.3 & 75.6 & 85.0 \\
SAFE & 49.5 & 49.5 & 53.0 & 49.0 & 50.2 & 52.2 & 51.9 & 50.0 & 50.0 & 49.8 & 49.7 & 49.8 & 50.1 & 50.0 & 50.0 & 49.8 & 50.3 & 50.3 \\
DRCT & 79.9 & 89.2 & 89.2 & 85.5 & 88.6 & 81.4 & 91.0 & 93.8 & 71.1 & 88.3 & 91.4 & 91.0 & 53.0 & 62.7 & 63.8 & 73.9 & 90.8 & 81.4 \\
AIDE & 52.9 & 51.1 & 60.2 & 49.8 & 69.3 & 70.1 & 93.6 & 60.6 & 89.0 & 49.6 & 51.6 & 51.0 & 72.1 & 66.5 & 59.0 & 80.6 & 54.5 & 63.6 \\
C2P-CLIP & 71.6 & 52.3 & 73.5 & 56.6 & 73.7 & 98.4 & 96.8 & 98.8 & \underline{99.3} & 62.3 & 77.5 & 76.9 & \textbf{99.6} & \underline{93.1} & 79.4 & \textbf{94.8} & 79.4 & 81.4 \\
AlignedForensics & 51.6 & 52.0 & 55.6 & 96.2 & 72.1 & 51.2 & 49.5 & 50.8 & 50.7 & 95.1 & \textbf{99.7} & \textbf{99.6} & 53.8 & 52.7 & 51.6 & 50.0 & \textbf{99.6} & 66.6 \\
DDA & \underline{89.5} & 94.6 & 89.6 & 95.6 & 76.6 & 91.0 & 72.5 & 92.7 & 92.8 & \underline{99.4} & 98.7 & 98.6 & 72.7 & 87.8 & \underline{90.2} & 52.1 & 98.8 & 87.8 \\
\midrule
Base & 70.1 & 88.5 & 73.2 & 66.5 & 71.5 & 90.7 & 60.0 & 99.0 & 87.4 & 97.2 & 99.1 & 99.0 & 49.4 & 80.2 & 75.9 & 81.0 & 99.1 & 81.6 \\
\rowcolor{tableblue} \textbf{Ours (MD-TTC)} & \textbf{89.7} & \underline{99.2} & \textbf{95.7} & \textbf{98.7} & \textbf{98.4} & 93.7 & 69.4 & 97.4 & 97.2 & 99.3 & 98.7 & 98.8 & 76.6 & 90.9 & 87.1 & \underline{91.9} & 98.8 & \textbf{93.0} \\
\rowcolor{tableblue} \textbf{Ours (GS-TTC)} & 84.0 & \textbf{99.4} & \underline{92.6} & \underline{96.7} & \underline{98.1} & \underline{98.9} & 72.4 & \textbf{99.4} & \textbf{99.4} & \textbf{99.6} & \underline{99.3} & \underline{99.5} & 62.0 & 86.4 & 85.1 & 81.9 & \underline{99.5} & \underline{91.4} \\
\bottomrule
\end{tabular}
\end{adjustbox}
\caption{Comparison on the AIGCDetect Benchmark under different adaptation protocols. Ours (MD-TTC) denotes the main mixed-domain protocol, and Ours (GS-TTC) denotes generator-specific adaptation. We report balanced accuracy (\%) for each generator subset and the average across subsets.}
\label{tab:persubset_aigcdetection}
\end{table*}
\begin{table*}[t]
\centering
\begin{adjustbox}{max width=\textwidth}
\begin{tabular}{lccccccccccl}
\toprule
Method & Janus & J-Pro-1B & J-Pro-7B & Show-o & LlamaGen & Infinity & VAR & PixArt-XL & SD3.5-L & FLUX & Avg. \\
\midrule
UnivFD & 70.6 & 68.3 & 59.4 & 57.9 & 75.2 & 56.7 & 66.7 & 53.9 & 74.0 & 51.6 & 63.4 \\
NPR & 30.1 & 29.8 & 29.8 & 79.6 & 79.5 & 79.8 & 79.3 & 78.7 & 79.4 & 79.1 & 64.5 \\
FatFormer & 48.8 & 48.6 & 48.8 & 73.2 & 84.3 & 53.9 & 90.1 & 67.1 & 71.0 & 54.6 & 64.0 \\
SAFE & 49.9 & 50.0 & 50.0 & \textbf{99.7} & 99.4 & \textbf{99.9} & 95.3 & \textbf{99.7} & 95.7 & \textbf{99.4} & 83.9\\
DRCT & 71.1 & 71.7 & 70.9 & 70.0 & 70.8 & 71.3 & 61.2 & 70.8 & 69.6 & 67.2 & 69.5 \\
AIDE & 50.6 & 52.6 & 51.5 & 98.7 & \textbf{99.7} & \underline{99.7} & 98.6 & 98.8 & \underline{99.0} & 95.4 & 84.5 \\
C2P-CLIP & 72.4 & 86.9 & 86.0 & 99.0 & 98.8 & 99.1 & 94.0 & 98.9 & 98.6 & 96.1 & 93.0 \\
AlignedForensics & 93.9 & 92.3 & 92.4 & 81.3 & 94.0 & 94.9 & 89.3 & 95.1 & 84.9 & 92.0 & 91.0  \\
DDA & 98.5 & \underline{99.3} & 98.7 & 94.8 & 99.3 & 98.9 & 80.4 & 99.4 & 97.0 & 97.2 & 96.3 \\
\midrule
Base & 95.5 & 98.8 & 97.3 & 95.4 & 98.0 & 64.4 & 84.1 & 94.6 & 50.5 &  50.6 & 82.9 \\
\rowcolor{tableblue} \textbf{Ours (MD-TTC)} & \underline{99.2} & \underline{99.3} & \underline{99.1} & \underline{99.1} & 99.2 & 99.2 & \underline{99.0} & 99.0 & \textbf{99.2} & \underline{99.1} & \textbf{99.1} \\
\rowcolor{tableblue} \textbf{Ours (GS-TTC)} & \textbf{99.4} & \textbf{99.5} & \textbf{99.4} & \textbf{99.7} & \textbf{99.9} & 99.1 & \textbf{99.6} & \underline{99.5} & 98.3 & 94.6 & \underline{98.9} \\
\bottomrule
\end{tabular}
\end{adjustbox}
\caption{Comparison on the AIGI-Holmes Benchmark under different adaptation protocols. Ours (MD-TTC) denotes the main mixed-domain protocol, and Ours (GS-TTC) denotes generator-specific adaptation. We report balanced accuracy (\%) for each generator subset and the average across subsets.}
\label{tab:persubset_aigi}
\end{table*}

\subsection{Comparison with Generator-Specific TTC}
The main experiments use mixed-domain TTC (MD-TTC), where adaptation is performed once on the full unlabeled test split without generator identities. Here we compare it with generator-specific TTC (GS-TTC), where TTC is adapted independently on each generator subset, to study whether separating target domains by generator is beneficial.

Tables~\ref{tab:persubset_aigcdetection} and~\ref{tab:persubset_aigi} show that MD-TTC achieves higher average performance than GS-TTC on both benchmarks: 93.0\% vs. 91.4\% on AIGCDetect, and 99.1\% vs. 98.9\% on AIGI-Holmes. This suggests that mixed-domain adaptation can exploit artifacts shared across generators through reliable pseudo labels from the full target set. However, this shared supervision also introduces a trade-off: it may dilute generator-specific cues and cause negative transfer on some subsets. For example, on AIGCDetect, MD-TTC slightly degrades from the Base detector on GauGAN, SD14, SD15, and Wukong, whereas GS-TTC avoids these drops and improves all subsets in these tables over the Base detector. This contrast shows that MD-TTC favors shared cross-generator cues, while GS-TTC better preserves generator-dependent evidence.

AIGI-Holmes further illustrates this trade-off under a more diverse generator composition. Rather than uniformly benefiting all subsets, the mixed target set makes the shared-versus-specific cue trade-off more apparent: MD-TTC is slightly worse than GS-TTC on several subsets, but substantially better on difficult ones such as SD3.5-L and FLUX. This suggests that cross-generator supervision helps challenging subsets borrow useful cues from related generators, while generator-specific adaptation better preserves isolated domain-specific evidence.

Overall, these results support our main protocol: TTC does not require generator identities and can benefit from shared artifacts in mixed-generator test distributions, while generator-specific adaptation provides a useful diagnostic view of the trade-off between shared and generator-specific cues.

\begin{table}[t]
\centering
\small
\setlength{\tabcolsep}{12pt}
\begin{tabular}{lc}
\toprule
Method & BAcc (\%) \\
\midrule
Base  & 81.6 \\
\midrule
TENT  & 50.0 \\
EATA  & 65.4 \\
SAR   & 50.0 \\
CoTTA & 83.8 \\
T3A   & \underline{84.4} \\
\midrule
TTC (Ours) & \textbf{93.0} \\
\bottomrule
\end{tabular}
\caption{
Comparison with general test-time adaptation methods on AIGCDetect Benchmark.
All methods are applied to the same pretrained Base detector and evaluated under the same mixed-domain test-time adaptation protocol.
}
\label{tab:tta_baselines}
\end{table}

\subsection{Comparison with General TTA methods.}
We further compare TTC with representative test-time adaptation methods under the same mixed-domain protocol on AIGCDetect Benchmark. As shown in Table~\ref{tab:tta_baselines}, TENT and SAR collapse to near-random
performance, while EATA also substantially degrades the Base detector. CoTTA and T3A provide moderate improvements, reaching 83.8\% and 84.4\% BAcc, respectively. In comparison, TTC achieves 93.0\%, outperforming the strongest general TTA baseline by 8.6 percentage points. 
Under the evaluated Base detector and mixed-domain protocol, the general-purpose TTA methods provide limited or unstable improvements, whereas TTC achieves a substantially higher BAcc. These results show that TTC is more effective in this setting by combining reliable, class-balanced pseudo-label selection with progressive curriculum adaptation.

\subsection{Generalization to Different Detector Backbones}
To demonstrate the model-agnostic nature of TTC, we apply it to two additional detector backbones, C2P-CLIP and AIDE. We evaluate these variants on two representative generator-shift benchmarks, which cover diverse generator families including GAN-, diffusion-, and autoregressive-based models. As shown in Table~\ref{tab:backbone}, TTC consistently improves both corresponding Base detectors. With AIDE, TTC improves balanced accuracy from 72.0\% to 89.0\% on AIGCDetect and from 73.3\% to 99.5\% on AIGI-Holmes. With C2P-CLIP, TTC further increases performance from 74.0\% to 93.2\% and from 87.5\% to 99.6\%, respectively. Notably, both TTC variants outperform DDA, the strongest baseline in Table~\ref{tab:aigcdetection} and Table~\ref{tab:aigi}, on both benchmarks. These results support the model-agnostic design of TTC, showing that the same adaptation protocol can effectively improve detectors built on different backbones across diverse generator distributions.

\begin{table}[t]
\centering
    \begin{adjustbox}{max width=\linewidth}
    \begin{tabular}{lcc}
    \toprule
    Method & AIGCDetect & AIGI-Holmes \\
    \midrule
    DDA~(reference) & 87.8 & 96.3 \\
    \midrule
    AIDE Base & 72.0 & 73.3 \\
    \rowcolor{tableblue} \textbf{TTC w/ AIDE} & \underline{89.0} & \underline{99.5} \\
    C2P-CLIP Base & 74.0 & 87.5 \\
    \rowcolor{tableblue} \textbf{TTC w/ C2P-CLIP} & \textbf{93.2} & \textbf{99.6} \\

\bottomrule
    \end{tabular}
    \end{adjustbox}
    \caption{Generalization of TTC to additional detector backbones, including C2P-CLIP and AIDE. DDA is included as a strong reference baseline.}
    \label{tab:backbone}
\end{table}

\subsection{Adaptation Data Efficiency.}
To examine whether TTC requires access to the full test set, we randomly sample $K$ unlabeled images from the mixed AIGCDetect Benchmark, using a fixed seed and perform TTC adaptation only on this sampled subset. Each setting is independently initialized from the same Base detector, while performance is evaluated on both the full test set and the non-overlapping remainder excluding all adaptation samples. Runtime is measured as GPU hours on a single NVIDIA H800 card.

As shown in Table~\ref{tab:adaptation_efficiency}, performance is not monotonic in the number of adaptation images. Only 10K adaptation images improve BAcc from 81.60\% to 94.46\% in 0.79 GPU hours, while 30K images achieve the best result of 94.62\% in 2.15 GPU hours, compared with 11.16 GPU hours for full-set adaptation. This suggests that a larger heterogeneous target batch may introduce redundant or noisy pseudo labels. We therefore interpret this experiment as evidence that TTC does not require access to the full test set, rather than as evidence that adaptation performance necessarily scales with target-set size.
Evaluating only on samples not used for adaptation produces nearly identical results, with differences of at most 0.10 percentage points. This shows that the gains transfer beyond the adaptation samples and are not mainly caused by same-sample fitting. The 10K and 30K settings substantially reduce adaptation cost compared with full-set TTC.

\begin{table*}[t]
\centering
\small
\setlength{\tabcolsep}{4.5pt}
\begin{tabular}{lrrrrrrr}
\toprule
\makecell{Adaptation\\size}
& Coverage
& Base
& R1
& R2
& R3
& \makecell{Excl.\\adapt.}
& \makecell{Time\\(GPU h)} \\
\midrule
1K
& 0.7\%
& 81.60
& 86.91
& 86.00
& 88.93
& 88.93
& 0.18 \\
10K
& 6.6\%
& 81.60
& 91.33
& 93.03
& 94.46
& 94.42
& 0.79 \\
30K
& 19.7\%
& 81.60
& 92.13
& 93.54
& \textbf{94.62}
& \textbf{94.52}
& 2.15 \\
76.25K
& 50.0\%
& 81.60
& 91.69
& 92.58
& 93.00
& 92.91
& 5.27 \\
152.598K (Full)
& 100.0\%
& 81.60
& 91.52
& 92.49
& 93.59
& --
& 11.16 \\
\bottomrule
\end{tabular}
\caption{
Adaptation data efficiency on AIGCDetect Benchmark.
TTC is adapted using $K$ unlabeled target images and evaluated on the full test set. ``Excl.\ adapt.'' reports R3 performance after removing all adaptation samples from evaluation.
}
\label{tab:adaptation_efficiency}
\end{table*}

\subsection{Stability Across Multiple Runs}
To evaluate the run-to-run stability of TTC, we repeat the test-time adaptation process three times under the same mixed-domain protocol. Table~\ref{tab:stability} reports the mean and standard deviation of balanced accuracy across runs. TTC exhibits small variance on all evaluated benchmarks, achieving \(93.0 \pm 0.42\) on AIGCDetect, \(99.1 \pm 0.15\) on AIGI-Holmes, and \(91.1 \pm 0.31\) on Chameleon. These results indicate that TTC is stable across independent adaptation runs and is not sensitive to a particular random seed or stochastic sampling order.

\begin{table}[t]
\centering
\begin{adjustbox}{max width=\linewidth}
\begin{tabular}{lc}
\toprule
Benchmark & BAcc (\%) \\
\midrule
AIGCDetect & \(93.0 \pm 0.42\) \\
AIGI-Holmes & \(99.1 \pm 0.15\) \\
Chameleon & \(91.1 \pm 0.31\) \\
\bottomrule
\end{tabular}
\end{adjustbox}
\caption{Stability of TTC across multiple (3) adaptation runs. We repeat TTC three times under the same mixed-domain protocol and report the mean and standard deviation of balanced accuracy (\%) across runs.}
\label{tab:stability}
\end{table}

\begin{figure}[t]
    \centering
    \includegraphics[width=\linewidth]{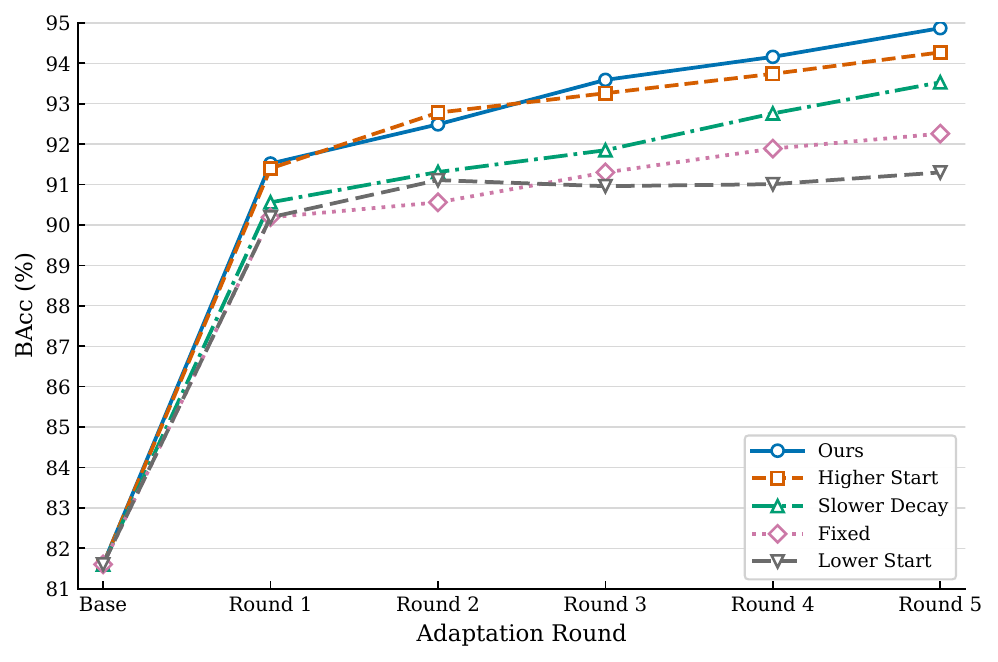}
    \caption{
    Sensitivity to curriculum schedules on AIGCDetect Benchmark.
    All schedules start from the Base detector. The default schedule (Ours) achieves the best final performance, while a higher starting threshold, slower decay, a fixed threshold, or a lower starting threshold
    leads to weaker adaptation.
    }
    \label{fig:schedule_sensitivity}
\end{figure}

\begin{figure}[t]
    \centering
    \includegraphics[width=\linewidth]{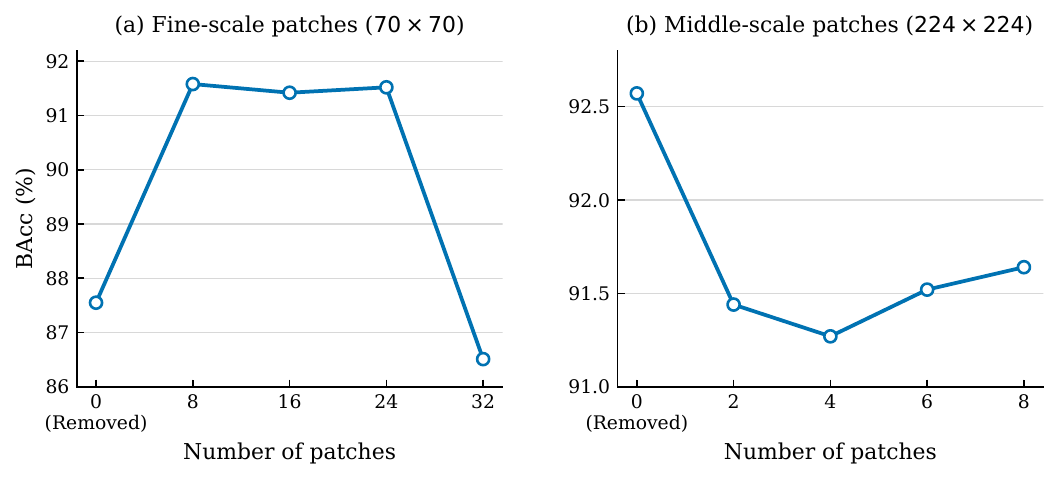}
    \caption{
    Sensitivity to patch numbers on AIGCDetect.
    We vary the number of fine-scale ($70{\times}70$) and middle-scale ($224{\times}224$) patches while keeping the remaining scales fixed. BAcc is reported  after the first adaptation round.
    }
    \label{fig:patch_sensitivity}
\end{figure}

\begin{figure}[t]
    \centering
    \includegraphics[width=\linewidth]{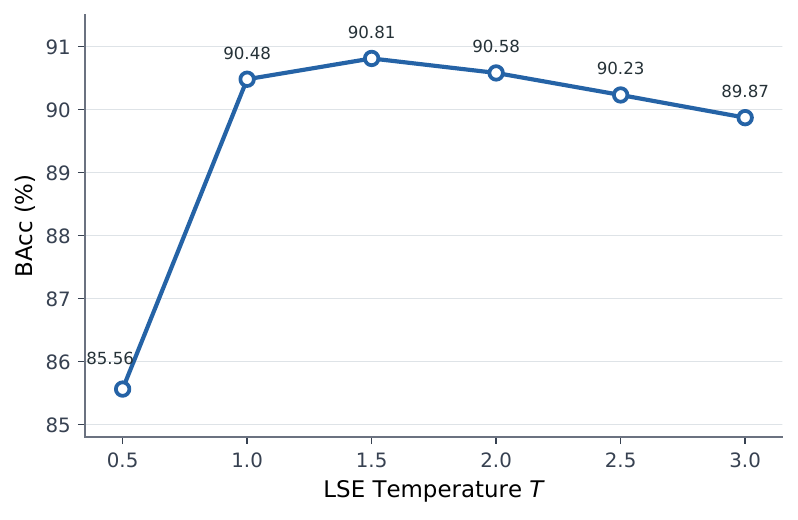}
    \caption{
    Sensitivity to the LSE temperature $T$ on AIGCDetect.
    We vary the $T$ while keeping all other settings fixed. BAcc is reported after the first adaptation round.
    }
    \label{fig:lse_temperature}
\end{figure}

\begin{table}[t]
\centering
\small
\setlength{\tabcolsep}{8pt}
\begin{tabular}{lc}
\toprule
Patch selection strategy & BAcc (\%) \\
\midrule
Random             & \textbf{91.16} \\
Score-only         & 90.07 \\
Score + diversity  & 90.43 \\
Full               & 90.55 \\
\bottomrule
\end{tabular}
\caption{
Sensitivity to patch selection strategies on AIGCDetect.
All variants use the same patch size and patch budget. BAcc is reported after the first adaptation round.
}
\label{tab:patch_selection}
\end{table}

\subsection{Hyperparameter Sensitivity Analysis}
\paragraph{Sensitivity to curriculum schedules.}
We evaluate five threshold schedules over five adaptation rounds to study the effect of the initial reliability requirement and the rate of curriculum relaxation. For this extended five-round analysis, we compare several representative relaxation schedules. This experiment is separate from the default three-round configuration used in the main experiments.

As shown in Fig.~\ref{fig:schedule_sensitivity}, all schedules start from the Base detector and produce substantial gains after the first round. Among the evaluated schedules, Ours achieves the best final BAcc of 94.87\%. Starting from a higher threshold yields comparable early performance but a lower final result, while slower decay leads to more conservative adaptation. The fixed and lower-start schedules perform substantially worse, showing that overly permissive early selection or the absence of progressive relaxation limits adaptation.

Specifically, Ours uses $(0.8,0.7,0.6,0.55,0.5)$; Higher Start uses $(0.9,0.8,0.7,0.6,0.5)$; Slower Decay uses $(0.8,0.75,0.7,0.65,0.6)$; Fixed uses $0.7$ throughout; and Lower Start uses $(0.7,0.65,0.6,0.55,0.5)$.

\paragraph{Sensitivity to patch numbers.}
We vary the number of patches at one scale while fixing the other scales to study the sensitivity of TTC to the multi-scale patch budget. 

As shown in Fig.~\ref{fig:patch_sensitivity}(a), removing the fine-scale ($70{\times}70$) patches substantially reduces the first-round performance, whereas using 8--24 patches yields stable results. Increasing the number to 32 leads to a clear degradation, suggesting that excessive local patches may introduce redundant or noisy supervision. 
In contrast, Fig.~\ref{fig:patch_sensitivity}(b) shows only modest variation across different numbers of middle-scale ($224{\times}224$) patches, indicating that TTC is relatively insensitive to the middle-scale patch budget. Overall, the method remains stable over a broad range of practical patch configurations without requiring precise tuning.

\paragraph{Sensitivity to LSE temperature.}
We further study the effect of the LSE temperature $T$ used in multi-scale prediction aggregation. As shown in Fig.~\ref{fig:lse_temperature}, performance remains stable for $T\in[1.0,2.5]$, where BAcc varies by only 0.58 percentage points. 
The setting $T=2.0$ achieves 90.58\% BAcc, only 0.23 points below the best result in this range. 
In contrast, an overly low temperature of $T=0.5$ causes a clear degradation, while larger temperatures lead to only a mild decline. These results indicate that the method is robust to the choice of $T$ over a broad practical range and does not rely on precise temperature tuning.

\paragraph{Sensitivity to patch selection strategies.}
We compare four patch selection strategies under the same patch budget.

\textit{Random} uniformly samples crops from the image.
\textit{Score-only} ranks sliding-window candidates using a weighted combination of high-frequency energy, gradient magnitude, and local texture variation, and selects the top-ranked patches.
\textit{Score + diversity} additionally applies greedy spatial suppression to reduce redundant overlapping crops.
\textit{Full} selects most patches using the score-and-diversity strategy and fills the remaining slots with randomly sampled crops to maintain broader image coverage.

As shown in Table~\ref{tab:patch_selection}, all four strategies achieve comparable Round-1 performance, with BAcc ranging from 90.07\% to 91.16\%. Random selection performs slightly better in this setting, while the full strategy remains competitive with the guided variants. This result indicates that TTC is not strongly dependent on a particular patch selection strategy.

\begin{figure*}[t]
\centering
\includegraphics[width=0.83\textwidth]{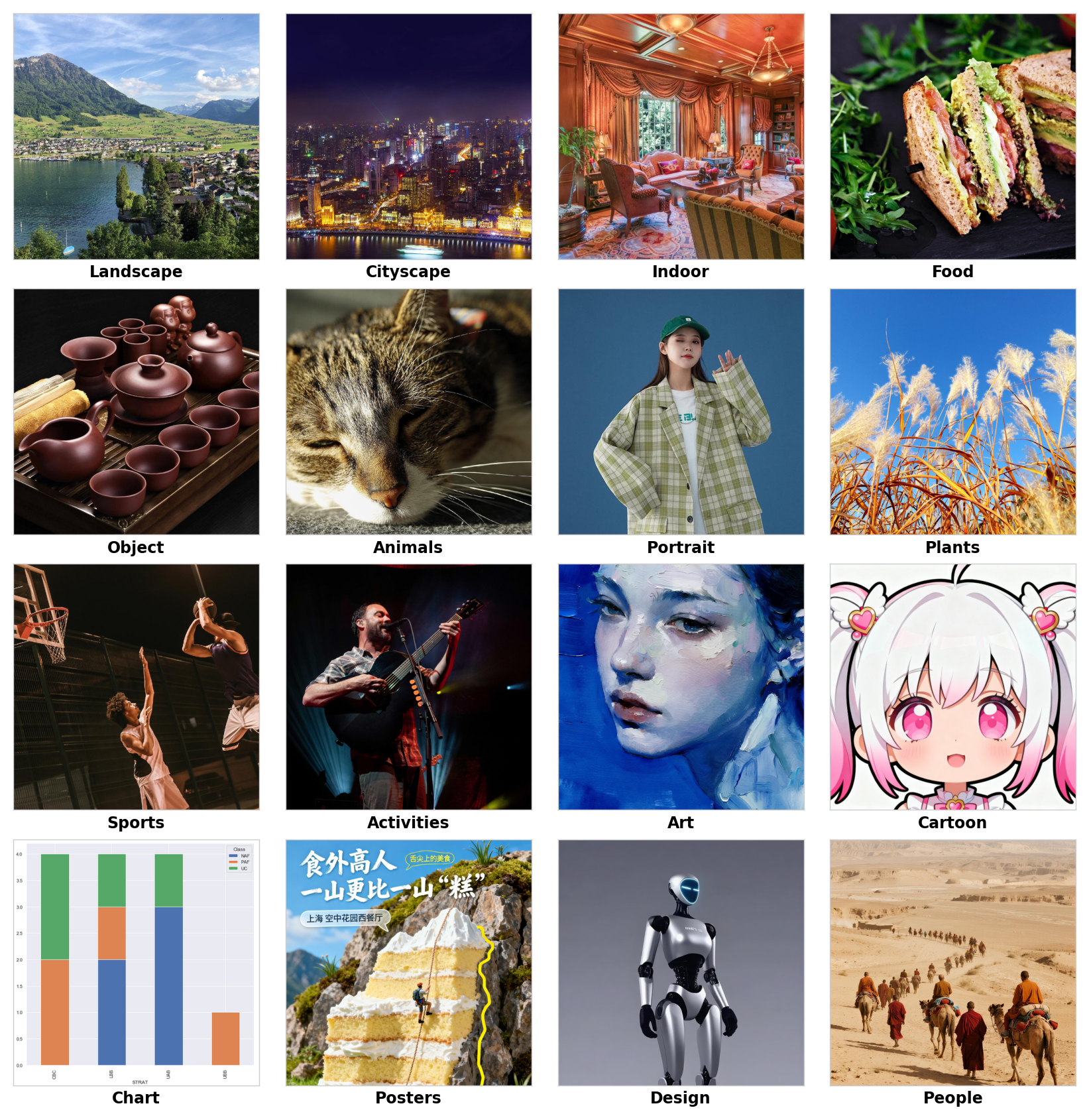}
\caption{Representative real images from 16 semantic categories in AIGCGuard.}
\label{fig:benchmark_real}
\end{figure*}

\begin{figure*}[t]
\centering
\includegraphics[width=\textwidth]{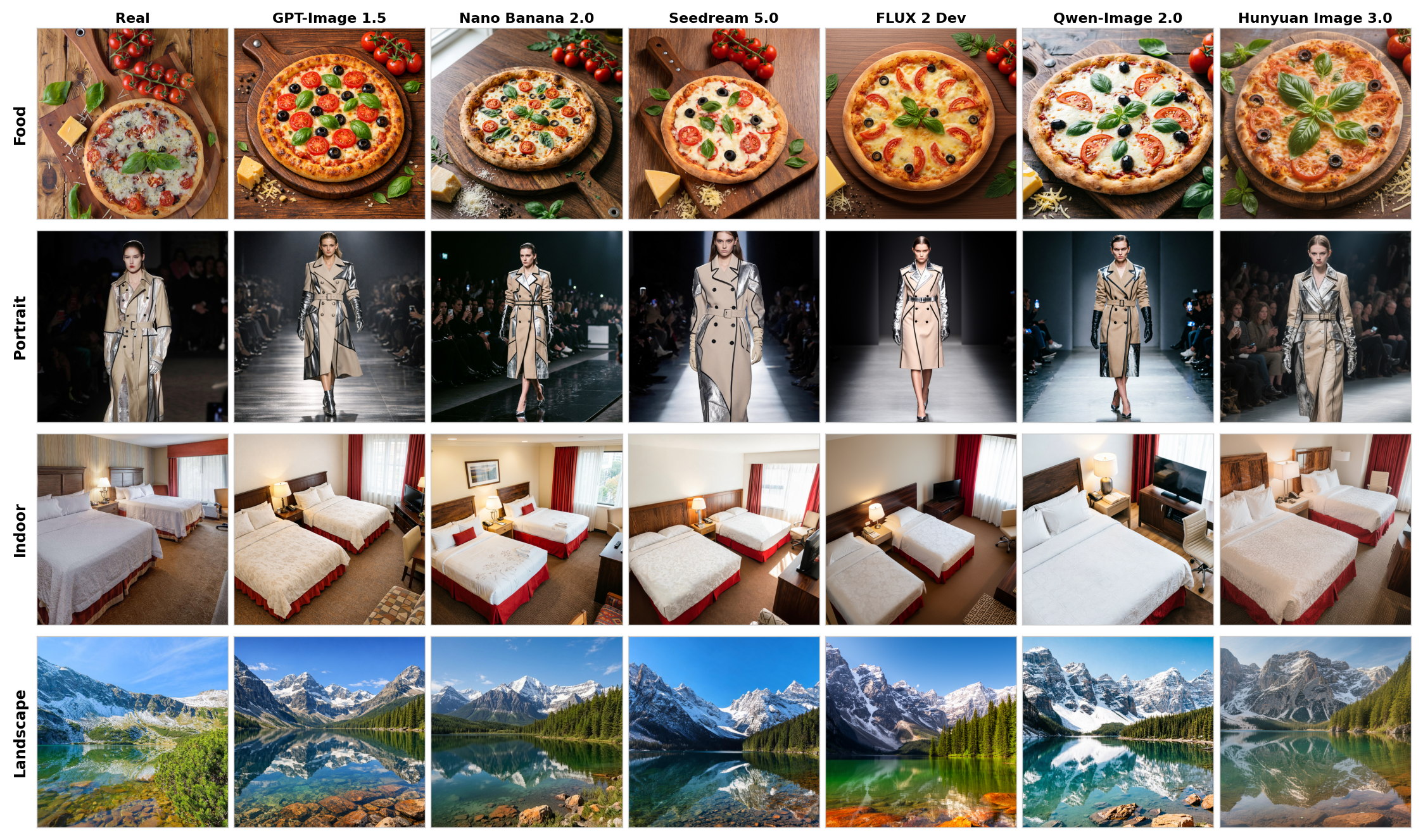}
\caption{Real images and samples generated by recent text-to-image models from the same prompts.}
\label{fig:benchmark_fake}
\end{figure*}

\subsection{Effect of Pseudo-label Quality.}
We analyze both naturally selected and synthetically corrupted pseudo labels to quantify when TTC remains stable and when erroneous supervision causes degradation.
Confidence alone cannot guarantee correctness under severe generator shift. TTC uses cross-scale consensus to improve pseudo-label reliability, while class-balanced selection, parameter anchoring, and a label-free safety check mitigate the impact of biased or erroneous supervision. We run the default three-round TTC and use ground-truth labels only for post-hoc evaluation of the selected pseudo labels.
As shown in Table~\ref{tab:pseudo_label_quality}, the selected pseudo labels maintain accuracies of 92.63\%, 91.69\%, and 92.55\% across the three rounds, and consistently improve performance across rounds.

We then conduct an oracle-controlled analysis to isolate the effect of pseudo-label accuracy. We fix the adaptation samples, synthetically construct pseudo labels at varying accuracy levels using ground-truth labels, and perform one adaptation round. TTC achieves 97.11\%, 95.10\%, 95.10\%, and 79.34\% BAcc with pseudo-label accuracies of 100\%, 95\%, 90\%, and 80\%, respectively. 
These results reveal a clear reliability regime: once pseudo-label accuracy reaches 90\%, further improvements in label accuracy have only a limited effect on the adaptation outcome. The selected pseudo labels of our TTC remain within this effective range across all three rounds. Performance degrades sharply below this regime, motivating the safety check and rollback mechanism.

\begin{table}[t]
\centering
\small
\setlength{\tabcolsep}{4.5pt}
\begin{tabular}{lcc}
\toprule
Setting & \makecell{Pseudo-label\\accuracy (\%)} & BAcc (\%) \\
\midrule
\multicolumn{3}{l}{\textit{(a) Our}} \\
Round 1 & 92.6 & 91.0 \\
Round 2 & 91.7 & 91.9 \\
Round 3 & 92.6 & 93.0 \\
\midrule
\multicolumn{3}{l}{\textit{(b) Oracle-controlled pseudo labels}} \\
100\% accuracy & 100 & 97.1 \\
95\% accuracy  & 95  & 95.1 \\
90\% accuracy  & 90  & 95.1 \\
85\% accuracy  & 85  & 89.9 \\
80\% accuracy  & 80  & 79.3 \\
\bottomrule
\end{tabular}
\caption{Effect of pseudo-label quality on AIGCDetect. Panel (a) reports TTC's three adaptation rounds; panel (b) fixes the adaptation samples and controls pseudo-label accuracy using ground-truth labels.}
\label{tab:pseudo_label_quality}
\end{table}

\subsection{Additional Implementation Details}

\noindent \textbf{Base Detector Training.}
Although TTC itself is applied only at test time, the Base detector is trained with lightweight image-level perturbations to improve robustness before adaptation. Specifically, JPEG compression, Gaussian blur, and resizing are randomly applied during source training. These perturbations are not used during TTC adaptation on clean benchmarks, except when explicitly evaluating robustness to perturbations. This separation keeps the reported clean benchmark results faithful to the original test distribution, while providing a more robust initialization for adaptation under natural distribution shifts.

\noindent \textbf{Patch Selection Details.}
For each local scale, candidate crops are extracted using a sliding window with a stride equal to half the patch size. Each candidate is scored using high-frequency energy, average gradient magnitude, and local intensity standard deviation, with weights $0.4$, $0.3$, and $0.3$, respectively.
Candidates are traversed in descending score order, and a crop is retained only if its Chebyshev distance from all previously selected crops is at least half the patch size. We set the random sampling ratio to $0.2$, and randomly sampled crops also fill any remaining slots when spatial suppression returns fewer candidates than required.

\noindent \textbf{Label-Free Safety Check}
Since ground-truth labels are unavailable during test-time adaptation, unreliable pseudo labels may cause the detector to drift or collapse. We therefore use a label-free safety check that compares each adapted curriculum round against the frozen Base detector using two prediction-only statistics: the predicted fake prior and the average prediction certainty. A round is considered unsafe in two cases. First, a large prior shift together with reduced certainty suggests that adaptation may be losing discriminability. Second, when the Base detector is already highly confident, a prior shift together with increased certainty may indicate overconfident pseudo-label confirmation. Because curriculum rounds are sequentially dependent, we use early stopping rather than treating rounds independently: once the first unsafe round is detected, TTC selects the previous round for evaluation, with the frozen Base detector serving as the fallback when no adapted round is safe. This prevents over-updating when pseudo labels are unreliable, while preserving adaptation on shifted target distributions. The safety check is applied uniformly across benchmarks using fixed thresholds and prediction-only statistics. The parameter-anchoring coefficient in Eq.~(6) is fixed to
$\lambda=0.001$ across all experiments.

\subsection{Details of Benchmark Curation}
\noindent \textbf{Principles.}
Our benchmark is designed to provide a high-quality, semantically diverse, and visually realistic dataset for evaluating AIGC detection methods. The construction emphasizes semantic coverage, intra-category diversity, and inclusion of advanced generative models, reflecting the current state of the most powerful text-to-image generative models. 

\noindent \textbf{Real Image Sampling.}
Real images are sampled from tens of billions of industrial Pre-Training and Mid-Training text-to-image datasets. To ensure semantic diversity, we assign images to 22 categories using the Qwen3-VL~\citep{qwen3} vision-language model, covering a broad range of real-world content such as objects, landscapes, animals, portraits, and charts. Within each category, we perform contrastive-learning-based K-means clustering~\citep{simclr,kmeans} to capture fine-grained semantic distinctions. From these clusters, a representative subset is selected via Facility Location selection~\citep{fl,coreset}, resulting in a representative subset of high-quality real images.

Selected images undergo quality control including image quality assessment, manual AIGC detection and NSFW filtering, and secondary caption re-annotation to ensure accurate textual descriptions. Additional manual inspection is performed to remove any remaining visually ambiguous or semantically inconsistent samples. After filtering and validation, the dataset contains 3,100 real images.

\noindent \textbf{Image Generation.}
Each real image is paired with counterparts generated by 40 state-of-the-art text-to-image models with generated image resolutions randomly varying from 512 pixels up to 4K, covering both open-source and proprietary systems: 
GPT-Image-2, GPT-Image-1.5, Nano banana, Seedream-4.0~\citep{seedream}, Seedream-5.0, Flux2-Dev, Flux2-Dev-Turbo, Qwen-Image-2.0, Qwen-Image-max~\citep{qwenimage}, Z-Image~\citep{zimage}, Z-Image-Turbo, Ernie-Image~\citep{ernie_image}, Ernie-Image-Turbo, Longcat-Image~\citep{longcatimage}, Hunyuan-Image-3.0~\citep{hunyuanimage}, Cosmos3~\citep{cosmos3}, GLM-Image, Omnigen2~\citep{omnigen2}, HiDream-o1~\citep{hidream}, Ideogram-4-fp8, openUni~\citep{openuni}, SRPO~\citep{SRPO}, FIBO~\citep{FIBO}, Infinity~\citep{infinity}, Kolors~\citep{kolors}, StableDiffusion3.5-Large, PixArt-Sigma~\citep{pixart}, Janus-Pro-7B~\citep{janus_pro}, BLIP3o~\citep{blip3o}, MMaDA~\citep{mmada}, BAGEL~\citep{bagel}, UniCoT~\citep{unicot}, Uni-Edit~\citep{uniedit}, Emu3.5~\citep{emu35}, Internvl-U~\citep{internvl}, Cheers~\citep{cheers}, Lance~\citep{lance}, STAR~\citep{star}, LatentUM~\citep{latentum}, and SenseNova-U1-8B-MoT~\citep{sensenova}.
Each model receives the same prompt independently, producing 40 generated images per real image, resulting in a total of 127,100 images. The dataset is balanced across generators, equalizing each generator’s contribution to the aggregate metric and providing a robust and high-quality benchmark for AIGC detection evaluation.

To reduce low-level shortcut cues arising from differences in image encoding and resolution, mild JPEG compression and resizing are incorporated into the construction of the AIGCGuard evaluation set.

\subsection{Visualization of Our Benchmark}
To qualitatively illustrate the diversity and difficulty of AIGCGuard, we provide two visualizations. Figure~\ref{fig:benchmark_real} shows representative real images from 16 semantic categories, covering natural scenes, city scenes, indoor environments, food, objects, animals, portraits, plants, sports, activities, people, artwork, cartoon-style images, posters, design content, and charts. These examples demonstrate the broad semantic coverage of the real-image split.

Figure~\ref{fig:benchmark_fake} further visualizes real images and generated samples from representative recent text-to-image models. Each row corresponds to the same prompt, with the first column showing the real image and the remaining columns showing generated counterparts. The generated samples preserve the high-level semantics of the real images while exhibiting model-specific variations in style, composition, and details. Their high visual fidelity makes AIGCGuard a challenging benchmark for open-set AIGC image detection.

\subsection{Limitations and Future Work}

TTC has several limitations. First, our primary protocol is offline and transductive: an unlabeled target batch is available during adaptation, and the current experiments do not establish online or streaming adaptation to future generator distributions. Second, TTC relies on the Base detector producing sufficiently reliable predictions for both classes. Systematic high-confidence errors, extreme target imbalance, or one-class prediction collapse may prevent useful adaptation, in which case TTC skips the corresponding round or falls back to the last safe checkpoint. Third, multi-scale prediction and iterative optimization introduce additional computation. Moreover, the noisy-or output is a heuristic decision score rather than a calibrated posterior probability. Finally, AIGCGuard complements in-the-wild evaluation with controlled, prompt-matched coverage of recent generators. Its generator-balanced construction is designed to isolate cross-generator detection performance rather than reproduce the naturally occurring real/fake prior or generator frequencies of a deployment stream. 


Future work will focus on more efficient and reliable test-time adaptation for AIGC detection, including reducing adaptation overhead, improving label-free reliability estimation, and designing safer update rules. Another important direction is streaming adaptation, where the detector must adapt to continuously changing distributions under limited memory and computation.
\bibliography{aaai2027}

@article{dda,
  title={Dual data alignment makes ai-generated image detector easier generalizable},
  author={Chen, Ruoxin and Xi, Junwei and Yan, Zhiyuan and Zhang, Ke-Yue and Wu, Shuang and Xie, Jingyi and Chen, Xu and Xu, Lei and Guan, Isabel and Yao, Taiping and others},
  journal={arXiv preprint arXiv:2505.14359},
  year={2025}
}

@article{seedream,
  title={Seedream 4.0: Toward next-generation multimodal image generation},
  author={Seedream, Team and Chen, Yunpeng and Gao, Yu and Gong, Lixue and Guo, Meng and Guo, Qiushan and Guo, Zhiyao and Hou, Xiaoxia and Huang, Weilin and Huang, Yixuan and others},
  journal={arXiv preprint arXiv:2509.20427},
  year={2025}
}

@article{zimage,
  title={Z-image: An efficient image generation foundation model with single-stream diffusion transformer},
  author={Cai, Huanqia and Cao, Sihan and Du, Ruoyi and Gao, Peng and Hoi, Steven and Hou, Zhaohui and Huang, Shijie and Jiang, Dengyang and Jin, Xin and Li, Liangchen and others},
  journal={arXiv preprint arXiv:2511.22699},
  year={2025}
}

@article{hunyuanimage,
  title={Hunyuanimage 3.0 technical report},
  author={Cao, Siyu and Chen, Hangting and Chen, Peng and Cheng, Yiji and Cui, Yutao and Deng, Xinchi and Dong, Ying and Gong, Kipper and Gu, Tianpeng and Gu, Xiusen and others},
  journal={arXiv preprint arXiv:2509.23951},
  year={2025}
}

@article{longcatimage,
  title={Longcat-image technical report},
  author={Team, Meituan LongCat and Ma, Hanghang and Tan, Haoxian and Huang, Jiale and Wu, Junqiang and He, Jun-Yan and Gao, Lishuai and Xiao, Songlin and Wei, Xiaoming and Ma, Xiaoqi and others},
  journal={arXiv preprint arXiv:2512.07584},
  year={2025}
}

@article{ernie5,
  title={ERNIE 5.0 Technical Report},
  author={Wang, Haifeng and Wu, Hua and Wu, Tian and Sun, Yu and Liu, Jing and Yu, Dianhai and Ma, Yanjun and He, Jingzhou and He, Zhongjun and Hong, Dou and others},
  journal={arXiv preprint arXiv:2602.04705},
  year={2026}
}

@article{qwenimage,
  title={Qwen-image technical report},
  author={Wu, Chenfei and Li, Jiahao and Zhou, Jingren and Lin, Junyang and Gao, Kaiyuan and Yan, Kun and Yin, Sheng-ming and Bai, Shuai and Xu, Xiao and Chen, Yilei and others},
  journal={arXiv preprint arXiv:2508.02324},
  year={2025}
}

@article{bnadapt,
  title={Improving robustness against common corruptions by covariate shift adaptation},
  author={Schneider, Steffen and Rusak, Evgenia and Eck, Luisa and Bringmann, Oliver and Brendel, Wieland and Bethge, Matthias},
  journal={Advances in neural information processing systems},
  volume={33},
  pages={11539--11551},
  year={2020}
}

@article{tbn,
  title={Evaluating prediction-time batch normalization for robustness under covariate shift},
  author={Nado, Zachary and Padhy, Shreyas and Sculley, D and D'Amour, Alexander and Lakshminarayanan, Balaji and Snoek, Jasper},
  journal={arXiv preprint arXiv:2006.10963},
  year={2020}
}

@article{t3a,
  title={Test-time classifier adjustment module for model-agnostic domain generalization},
  author={Iwasawa, Yusuke and Matsuo, Yutaka},
  journal={Advances in Neural Information Processing Systems},
  volume={34},
  pages={2427--2440},
  year={2021}
}

@article{delta,
  title={Delta: degradation-free fully test-time adaptation},
  author={Zhao, Bowen and Chen, Chen and Xia, Shu-Tao},
  journal={arXiv preprint arXiv:2301.13018},
  year={2023}
}

@article{tent,
  title={Tent: Fully test-time adaptation by entropy minimization},
  author={Wang, Dequan and Shelhamer, Evan and Liu, Shaoteng and Olshausen, Bruno and Darrell, Trevor},
  journal={arXiv preprint arXiv:2006.10726},
  year={2020}
}

@inproceedings{tda,
  title={Efficient test-time adaptation of vision-language models},
  author={Karmanov, Adilbek and Guan, Dayan and Lu, Shijian and El Saddik, Abdulmotaleb and Xing, Eric},
  booktitle={Proceedings of the IEEE/CVF Conference on Computer Vision and Pattern Recognition},
  pages={14162--14171},
  year={2024}
}

@inproceedings{eata,
  title={Efficient test-time model adaptation without forgetting},
  author={Niu, Shuaicheng and Wu, Jiaxiang and Zhang, Yifan and Chen, Yaofo and Zheng, Shijian and Zhao, Peilin and Tan, Mingkui},
  booktitle={International conference on machine learning},
  pages={16888--16905},
  year={2022},
  organization={PMLR}
}

@article{sar,
  title={Towards stable test-time adaptation in dynamic wild world},
  author={Niu, Shuaicheng and Wu, Jiaxiang and Zhang, Yifan and Wen, Zhiquan and Chen, Yaofo and Zhao, Peilin and Tan, Mingkui},
  journal={arXiv preprint arXiv:2302.12400},
  year={2023}
}

@article{tta_survey,
  title={A comprehensive survey on test-time adaptation under distribution shifts},
  author={Liang, Jian and He, Ran and Tan, Tieniu},
  journal={International Journal of Computer Vision},
  volume={133},
  number={1},
  pages={31--64},
  year={2025},
  publisher={Springer}
}

@inproceedings{robust_tta,
  title={Robust test-time adaptation in dynamic scenarios},
  author={Yuan, Longhui and Xie, Binhui and Li, Shuang},
  booktitle={Proceedings of the IEEE/CVF Conference on Computer Vision and Pattern Recognition},
  pages={15922--15932},
  year={2023}
}

@inproceedings{cotta,
  title={Continual test-time domain adaptation},
  author={Wang, Qin and Fink, Olga and Van Gool, Luc and Dai, Dengxin},
  booktitle={Proceedings of the IEEE/CVF conference on computer vision and pattern recognition},
  pages={7201--7211},
  year={2022}
}

@article{aigc_survey,
  title={Survey for detecting AI-generated content},
  author={Wang, Yu},
  journal={Advances in Engineering Technology Research},
  volume={11},
  number={1},
  pages={643--643},
  year={2024}
}

@inproceedings{npr,
  title={Rethinking the up-sampling operations in cnn-based generative network for generalizable deepfake detection},
  author={Tan, Chuangchuang and Zhao, Yao and Wei, Shikui and Gu, Guanghua and Liu, Ping and Wei, Yunchao},
  booktitle={Proceedings of the IEEE/CVF conference on computer vision and pattern recognition},
  pages={28130--28139},
  year={2024}
}

@inproceedings{univfd,
  title={Towards universal fake image detectors that generalize across generative models},
  author={Ojha, Utkarsh and Li, Yuheng and Lee, Yong Jae},
  booktitle={Proceedings of the IEEE/CVF conference on computer vision and pattern recognition},
  pages={24480--24489},
  year={2023}
}

@inproceedings{fatformer,
  title={Forgery-aware adaptive transformer for generalizable synthetic image detection},
  author={Liu, Huan and Tan, Zichang and Tan, Chuangchuang and Wei, Yunchao and Wang, Jingdong and Zhao, Yao},
  booktitle={Proceedings of the IEEE/CVF Conference on Computer Vision and Pattern Recognition},
  pages={10770--10780},
  year={2024}
}

@inproceedings{c2pclip,
  title={C2p-clip: Injecting category common prompt in clip to enhance generalization in deepfake detection},
  author={Tan, Chuangchuang and Tao, Renshuai and Liu, Huan and Gu, Guanghua and Wu, Baoyuan and Zhao, Yao and Wei, Yunchao},
  booktitle={Proceedings of the AAAI Conference on Artificial Intelligence},
  volume={39},
  number={7},
  pages={7184--7192},
  year={2025}
}

@inproceedings{aigihlomes,
  title={Aigi-holmes: Towards explainable and generalizable ai-generated image detection via multimodal large language models},
  author={Zhou, Ziyin and Luo, Yunpeng and Wu, Yuanchen and Sun, Ke and Ji, Jiayi and Yan, Ke and Ding, Shouhong and Sun, Xiaoshuai and Wu, Yunsheng and Ji, Rongrong},
  booktitle={Proceedings of the IEEE/CVF International Conference on Computer Vision},
  pages={18746--18758},
  year={2025}
}

@inproceedings{safe,
  title={Improving synthetic image detection towards generalization: An image transformation perspective},
  author={Li, Ouxiang and Cai, Jiayin and Hao, Yanbin and Jiang, Xiaolong and Hu, Yao and Feng, Fuli},
  booktitle={Proceedings of the 31st ACM SIGKDD Conference on Knowledge Discovery and Data Mining V. 1},
  pages={2405--2414},
  year={2025}
}

@inproceedings{drct,
  title={Drct: Diffusion reconstruction contrastive training towards universal detection of diffusion generated images},
  author={Chen, Baoying and Zeng, Jishen and Yang, Jianquan and Yang, Rui},
  booktitle={Forty-first International Conference on Machine Learning},
  year={2024}
}

@article{alignedforensics,
  title={Aligned datasets improve detection of latent diffusion-generated images},
  author={Rajan, Anirudh Sundara and Ojha, Utkarsh and Schloesser, Jedidiah and Lee, Yong Jae},
  journal={arXiv preprint arXiv:2410.11835},
  year={2024}
}

@article{aide,
  title={A sanity check for ai-generated image detection},
  author={Yan, Shilin and Li, Ouxiang and Cai, Jiayin and Hao, Yanbin and Jiang, Xiaolong and Hu, Yao and Xie, Weidi},
  journal={arXiv preprint arXiv:2406.19435},
  year={2024}
}

@article{patchcraft,
  title={Rich and poor texture contrast: A simple yet effective approach for ai-generated image detection},
  author={Zhong, Nan and Xu, Yiran and Qian, Zhenxing and Zhang, Xinpeng},
  journal={arXiv preprint arXiv:2311.12397},
  volume={3},
  number={6},
  pages={1},
  year={2023}
}

@article{aigccolor,
  title={Detecting gan-generated imagery using color cues},
  author={McCloskey, Scott and Albright, Michael},
  journal={arXiv preprint arXiv:1812.08247},
  year={2018}
}

@inproceedings{aigcdetect,
  title={CNN-generated images are surprisingly easy to spot... for now},
  author={Wang, Sheng-Yu and Wang, Oliver and Zhang, Richard and Owens, Andrew and Efros, Alexei A},
  booktitle={Proceedings of the IEEE/CVF conference on computer vision and pattern recognition},
  pages={8695--8704},
  year={2020}
}

@article{ttasafe,
  title={Towards stable test-time adaptation in dynamic wild world},
  author={Niu, Shuaicheng and Wu, Jiaxiang and Zhang, Yifan and Wen, Zhiquan and Chen, Yaofo and Zhao, Peilin and Tan, Mingkui},
  journal={arXiv preprint arXiv:2302.12400},
  year={2023}
}

@inproceedings{ttalabel,
  title={Label shift adapter for test-time adaptation under covariate and label shifts},
  author={Park, Sunghyun and Yang, Seunghan and Choo, Jaegul and Yun, Sungrack},
  booktitle={Proceedings of the IEEE/CVF International Conference on Computer Vision},
  pages={16421--16431},
  year={2023}
}

@article{ttn,
  title={Ttn: A domain-shift aware batch normalization in test-time adaptation},
  author={Lim, Hyesu and Kim, Byeonggeun and Choo, Jaegul and Choi, Sungha},
  journal={arXiv preprint arXiv:2302.05155},
  year={2023}
}

@article{aigc_co,
  title={Detecting GAN generated fake images using co-occurrence matrices},
  author={Nataraj, Lakshmanan and Mohammed, Tajuddin Manhar and Chandrasekaran, Shivkumar and Flenner, Arjuna and Bappy, Jawadul H and Roy-Chowdhury, Amit K and Manjunath, BS},
  journal={arXiv preprint arXiv:1903.06836},
  year={2019}
}

@article{genimage,
  title={Genimage: A million-scale benchmark for detecting ai-generated image},
  author={Zhu, Mingjian and Chen, Hanting and Yan, Qiangyu and Huang, Xudong and Lin, Guanyu and Li, Wei and Tu, Zhijun and Hu, Hailin and Hu, Jie and Wang, Yunhe},
  journal={Advances in neural information processing systems},
  volume={36},
  pages={77771--77782},
  year={2023}
}

@inproceedings{lare,
  title={Lare\^{} 2: Latent reconstruction error based method for diffusion-generated image detection},
  author={Luo, Yunpeng and Du, Junlong and Yan, Ke and Ding, Shouhong},
  booktitle={Proceedings of the IEEE/CVF Conference on Computer Vision and Pattern Recognition},
  pages={17006--17015},
  year={2024}
}

@inproceedings{aeroblade,
  title={Aeroblade: Training-free detection of latent diffusion images using autoencoder reconstruction error},
  author={Ricker, Jonas and Lukovnikov, Denis and Fischer, Asja},
  booktitle={Proceedings of the IEEE/CVF Conference on Computer Vision and Pattern Recognition},
  pages={9130--9140},
  year={2024}
}

@article{ssp,
  title={A single simple patch is all you need for ai-generated image detection},
  author={Chen, Jiaxuan and Yao, Jieteng and Niu, Li},
  journal={arXiv preprint arXiv:2402.01123},
  year={2024}
}

@inproceedings{bfree,
  title={A bias-free training paradigm for more general ai-generated image detection},
  author={Guillaro, Fabrizio and Zingarini, Giada and Usman, Ben and Sud, Avneesh and Cozzolino, Davide and Verdoliva, Luisa},
  booktitle={Proceedings of the Computer Vision and Pattern Recognition Conference},
  pages={18685--18694},
  year={2025}
}

@inproceedings{semgir,
  title={SemGIR: Semantic-guided image regeneration based method for AI-generated image detection and attribution},
  author={Yu, Xiao and Chen, Kejiang and Zeng, Kai and Fang, Han and Yang, Zijin and Shang, Xiuwei and Qi, Yuang and Zhang, Weiming and Yu, Nenghai},
  booktitle={Proceedings of the 32nd ACM International Conference on Multimedia},
  pages={8480--8488},
  year={2024}
}

@inproceedings{aigc_bias,
  title={Fake or jpeg? revealing common biases in generated image detection datasets},
  author={Grommelt, Patrick and Weiss, Louis and Pfreundt, Franz-Josef and Keuper, Janis},
  booktitle={European Conference on Computer Vision},
  pages={80--95},
  year={2024},
  organization={Springer}
}

@inproceedings{fakeinversion,
  title={Fakeinversion: Learning to detect images from unseen text-to-image models by inverting stable diffusion},
  author={Cazenavette, George and Sud, Avneesh and Leung, Thomas and Usman, Ben},
  booktitle={Proceedings of the IEEE/CVF Conference on Computer Vision and Pattern Recognition},
  pages={10759--10769},
  year={2024}
}

@inproceedings{fredect,
  title={Leveraging frequency analysis for deep fake image recognition},
  author={Frank, Joel and Eisenhofer, Thorsten and Sch{\"o}nherr, Lea and Fischer, Asja and Kolossa, Dorothea and Holz, Thorsten},
  booktitle={International conference on machine learning},
  pages={3247--3258},
  year={2020},
  organization={PMLR}
}

@article{curriculum_survey,
  title={A survey on curriculum learning},
  author={Wang, Xin and Chen, Yudong and Zhu, Wenwu},
  journal={IEEE transactions on pattern analysis and machine intelligence},
  volume={44},
  number={9},
  pages={4555--4576},
  year={2021},
  publisher={IEEE}
}

@article{bce,
  title={The regression analysis of binary sequences},
  author={Cox, David R},
  journal={Journal of the Royal Statistical Society Series B: Statistical Methodology},
  volume={20},
  number={2},
  pages={215--232},
  year={1958},
  publisher={Oxford University Press}
}

@inproceedings{bientropy,
  title={A sequential algorithm for training text classifiers: Corrigendum and additional data},
  author={Lewis, David D},
  booktitle={Acm sigir forum},
  volume={29},
  number={2},
  pages={13--19},
  year={1995},
  organization={ACM New York, NY, USA}
}

@article{l2,
  title={A simple weight decay can improve generalization},
  author={Krogh, Anders and Hertz, John},
  journal={Advances in neural information processing systems},
  volume={4},
  year={1991}
}

@article{cycleself,
  title={Cycle self-training for domain adaptation},
  author={Liu, Hong and Wang, Jianmin and Long, Mingsheng},
  journal={Advances in Neural Information Processing Systems},
  volume={34},
  pages={22968--22981},
  year={2021}
}

@article{semiself,
  title={Semi-supervised self-training of object detection models},
  author={Rosenberg, Chuck and Hebert, Martial and Schneiderman, Henry},
  year={2005},
  publisher={Carnegie Mellon University}
}

@article{selftraining,
  title={Self-training: A survey},
  author={Amini, Massih-Reza and Feofanov, Vasilii and Pauletto, Loic and Hadjadj, Lies and Devijver, Emilie and Maximov, Yury},
  journal={Neurocomputing},
  volume={616},
  pages={128904},
  year={2025},
  publisher={Elsevier}
}

@inproceedings{dire,
  title={Dire for diffusion-generated image detection},
  author={Wang, Zhendong and Bao, Jianmin and Zhou, Wengang and Wang, Weilun and Hu, Hezhen and Chen, Hong and Li, Houqiang},
  booktitle={Proceedings of the IEEE/CVF International Conference on Computer Vision},
  pages={22445--22455},
  year={2023}
}

@article{allpatch,
  title={All patches matter, more patches better: Enhance ai-generated image detection via panoptic patch learning},
  author={Yang, Zheng and Chen, Ruoxin and Yan, Zhiyuan and Zhang, Ke-Yue and Fu, Xinghe and Wu, Shuang and Shu, Xiujun and Yao, Taiping and Ding, Shouhong and Qin, Zequn and others},
  journal={arXiv preprint arXiv:2504.01396},
  year={2025}
}

@book{noisyor,
  title={Probabilistic reasoning in intelligent systems: networks of plausible inference},
  author={Pearl, Judea},
  year={2014},
  publisher={Elsevier}
}

@inproceedings{lse,
  title={Attention-based deep multiple instance learning},
  author={Ilse, Maximilian and Tomczak, Jakub and Welling, Max},
  booktitle={International conference on machine learning},
  pages={2127--2136},
  year={2018},
  organization={PMLR}
}

@article{mil_survey,
  title={A comprehensive review on multiple instance learning},
  author={Fatima, Samman and Ali, Sikandar and Kim, Hee-Cheol},
  journal={Electronics},
  volume={12},
  number={20},
  pages={4323},
  year={2023},
  publisher={MDPI}
}

@inproceedings{crossscale,
  title={Path aggregation network for instance segmentation},
  author={Liu, Shu and Qi, Lu and Qin, Haifang and Shi, Jianping and Jia, Jiaya},
  booktitle={Proceedings of the IEEE conference on computer vision and pattern recognition},
  pages={8759--8768},
  year={2018}
}

@inproceedings{multiscale,
  title={Feature pyramid networks for object detection},
  author={Lin, Tsung-Yi and Doll{\'a}r, Piotr and Girshick, Ross and He, Kaiming and Hariharan, Bharath and Belongie, Serge},
  booktitle={Proceedings of the IEEE conference on computer vision and pattern recognition},
  pages={2117--2125},
  year={2017}
}

@article{qwen3,
  title={Qwen3-vl technical report},
  author={Bai, Shuai and Cai, Yuxuan and Chen, Ruizhe and Chen, Keqin and Chen, Xionghui and Cheng, Zesen and Deng, Lianghao and Ding, Wei and Gao, Chang and Ge, Chunjiang and others},
  journal={arXiv preprint arXiv:2511.21631},
  year={2025}
}

@article{mean,
  title={Mean teachers are better role models: Weight-averaged consistency targets improve semi-supervised deep learning results},
  author={Tarvainen, Antti and Valpola, Harri},
  journal={Advances in neural information processing systems},
  volume={30},
  year={2017}
}

@inproceedings{simclr,
  title={A simple framework for contrastive learning of visual representations},
  author={Chen, Ting and Kornblith, Simon and Norouzi, Mohammad and Hinton, Geoffrey},
  booktitle={International conference on machine learning},
  pages={1597--1607},
  year={2020},
  organization={PmLR}
}

@inproceedings{kmeans,
  title={Deep clustering for unsupervised learning of visual features},
  author={Caron, Mathilde and Bojanowski, Piotr and Joulin, Armand and Douze, Matthijs},
  booktitle={Proceedings of the European conference on computer vision (ECCV)},
  pages={132--149},
  year={2018}
}

@article{coreset,
  title={Active learning for convolutional neural networks: A core-set approach},
  author={Sener, Ozan and Savarese, Silvio},
  journal={arXiv preprint arXiv:1708.00489},
  year={2017}
}

@inproceedings{fl,
  title={Submodularity in data subset selection and active learning},
  author={Wei, Kai and Iyer, Rishabh and Bilmes, Jeff},
  booktitle={International conference on machine learning},
  pages={1954--1963},
  year={2015},
  organization={PMLR}
}

@article{cosmos3,
  title={Cosmos 3: Omnimodal world models for physical ai},
  author={Agarwal, Niket and Ali, Arslan and Allen, Jon and Antolini, Martin and Aubame, Adeline and Azzolini, Alisson and Bai, Junjie and Bala, Maciej and Balaji, Yogesh and Bapst, Josh and others},
  journal={arXiv preprint arXiv:2606.02800},
  year={2026}
}

@article{omnigen2,
  title={Omnigen2: Exploration to advanced multimodal generation},
  author={Wu, Chenyuan and Zheng, Pengfei and Yan, Ruiran and Xiao, Shitao and Luo, Xin and Wang, Yueze and Li, Wanli and Jiang, Xiyan and Liu, Yexin and Zhou, Junjie and others},
  journal={arXiv preprint arXiv:2506.18871},
  year={2025}
}

@article{hidream,
  title={Hidream-o1-image: A natively unified image generative foundation model with pixel-level unified transformer},
  author={Cai, Qi and Chen, Jingwen and Gao, Chengmin and Gong, Zijian and Li, Yehao and Pan, Yingwei and Peng, Yi and Qiu, Zhaofan and Yu, Kai and Zhang, Yiheng and others},
  journal={arXiv preprint arXiv:2605.11061},
  year={2026}
}

@article{openuni,
  title={Openuni: A simple baseline for unified multimodal understanding and generation},
  author={Wu, Size and Wu, Zhonghua and Gong, Zerui and Tao, Qingyi and Jin, Sheng and Li, Qinyue and Li, Wei and Loy, Chen Change},
  journal={arXiv preprint arXiv:2505.23661},
  year={2025}
}

@article{FIBO,
  title={Generating an Image From 1,000 Words: Enhancing Text-to-Image With Structured Captions},
  author={Gutflaish, Eyal and Kachlon, Eliran and Zisman, Hezi and Hacham, Tal and Sarid, Nimrod and Visheratin, Alexander and Huberman, Saar and Davidi, Gal and Bukchin, Guy and Goldberg, Kfir and others},
  journal={arXiv preprint arXiv:2511.06876},
  year={2025}
}

@misc{SRPO,
      title={Directly Aligning the Full Diffusion Trajectory with Fine-Grained Human Preference}, 
      author={Xiangwei Shen and Zhimin Li and Zhantao Yang and Shiyi Zhang and Yingfang Zhang and Donghao Li and Chunyu Wang and Qinglin Lu and Yansong Tang},
      year={2025},
      eprint={2509.06942},
      archivePrefix={arXiv},
      primaryClass={cs.AI},
      url={https://arxiv.org/abs/2509.06942}, 
}

@inproceedings{infinity,
  title={Infinity: Scaling bitwise autoregressive modeling for high-resolution image synthesis},
  author={Han, Jian and Liu, Jinlai and Jiang, Yi and Yan, Bin and Zhang, Yuqi and Yuan, Zehuan and Peng, Bingyue and Liu, Xiaobing},
  booktitle={Proceedings of the Computer Vision and Pattern Recognition Conference},
  pages={15733--15744},
  year={2025}
}

@article{kolors,
  title={Kolors: Effective Training of Diffusion Model for Photorealistic Text-to-Image Synthesis},
  author={Kolors Team},
  journal={arXiv preprint},
  year={2024}
}

@article{bagel,
  title   = {Emerging Properties in Unified Multimodal Pretraining},
  author  = {Deng, Chaorui and Zhu, Deyao and Li, Kunchang and Gou, Chenhui and Li, Feng and Wang, Zeyu and Zhong, Shu and Yu, Weihao and Nie, Xiaonan and Song, Ziang and Shi, Guang and Fan, Haoqi},
  journal = {arXiv preprint arXiv:2505.14683},
  year    = {2025}
}

@inproceedings{pixart,
  title={Pixart-$\sigma$: Weak-to-strong training of diffusion transformer for 4k text-to-image generation},
  author={Chen, Junsong and Ge, Chongjian and Xie, Enze and Wu, Yue and Yao, Lewei and Ren, Xiaozhe and Wang, Zhongdao and Luo, Ping and Lu, Huchuan and Li, Zhenguo},
  booktitle={European Conference on Computer Vision},
  pages={74--91},
  year={2024},
  organization={Springer}
}

@article{janus_pro,
  title={Janus-pro: Unified multimodal understanding and generation with data and model scaling},
  author={Chen, Xiaokang and Wu, Zhiyu and Liu, Xingchao and Pan, Zizheng and Liu, Wen and Xie, Zhenda and Yu, Xingkai and Ruan, Chong},
  journal={arXiv preprint arXiv:2501.17811},
  year={2025}
}

@article{blip3o,
  title={Blip3-o: A family of fully open unified multimodal models-architecture, training and dataset},
  author={Chen, Jiuhai and Xu, Zhiyang and Pan, Xichen and Hu, Yushi and Qin, Can and Goldstein, Tom and Huang, Lifu and Zhou, Tianyi and Xie, Saining and Savarese, Silvio and others},
  journal={arXiv preprint arXiv:2505.09568},
  year={2025}
}

@article{mmada,
  title={Mmada: Multimodal large diffusion language models},
  author={Yang, Ling and Tian, Ye and Li, Bowen and Zhang, Xinchen and Shen, Ke and Tong, Yunhai and Wang, Mengdi},
  journal={Advances in Neural Information Processing Systems},
  volume={38},
  pages={138867--138907},
  year={2026}
}

@article{unicot,
  title={Uni-cot: Towards unified chain-of-thought reasoning across text and vision},
  author={Qin, Luozheng and Gong, Jia and Sun, Yuqing and Li, Tianjiao and Yang, Mengping and Yang, Xiaomeng and Qu, Chao and Tan, Zhiyu and Li, Hao},
  journal={arXiv preprint arXiv:2508.05606},
  year={2025}
}

@article{uniedit,
  title   = {Uni-Edit: Intelligent Editing Is A General Task For Unified Model Tuning},
  author  = {Zheng, Dian and Zhang, Manyuan and Li, Hongyu and Liu, Hongbo and Zou, Kai and Feng, Kaituo and Li, Hongsheng},
  journal = {arXiv preprint arXiv:2605.21487},
  year    = {2026}
}

@article{emu35,
  title={Emu3. 5: Native multimodal models are world learners},
  author={Cui, Yufeng and Chen, Honghao and Deng, Haoge and Huang, Xu and Li, Xinghang and Liu, Jirong and Liu, Yang and Luo, Zhuoyan and Wang, Jinsheng and Wang, Wenxuan and others},
  journal={arXiv preprint arXiv:2510.26583},
  year={2025}
}

@article{internvl,
  title={Internvl-u: Democratizing unified multimodal models for understanding, reasoning, generation and editing},
  author={Tian, Changyao and Yang, Danni and Chen, Guanzhou and Cui, Erfei and Wang, Zhaokai and Duan, Yuchen and Yin, Penghao and Chen, Sitao and Yang, Ganlin and Liu, Mingxin and others},
  journal={arXiv preprint arXiv:2603.09877},
  year={2026}
}

@article{cheers,
  title={Cheers: Decoupling Patch Details from Semantic Representations Enables Unified Multimodal Comprehension and Generation},
  author={Zhang, Yichen and Peng, Da and Guo, Zonghao and Zhang, Zijian and Yang, Xuesong and Sun, Tong and Sun, Shichu and Zhang, Yidan and Li, Yanghao and Zhao, Haiyan and others},
  journal={arXiv preprint arXiv:2603.12793},
  year={2026}
}

@misc{lance,
      title         = {Lance: Unified Multimodal Modeling by Multi-Task Synergy},
      author        = {Fengyi Fu and Mengqi Huang and Shaojin Wu and Yunsheng Jiang and Yufei Huo and Hao Li and Yinghang Song and Fei Ding and Jianzhu Guo and Qian He and Zheren Fu and Zhendong Mao and Yongdong Zhang},
      year          = {2026},
      eprint        = {2605.18678},
      archivePrefix = {arXiv},
      primaryClass  = {cs.CV},
      url           = {https://arxiv.org/abs/2605.18678},
}

@article{star,
  title={STAR: STacked AutoRegressive Scheme for Unified Multimodal Learning},
  author={Qin, Jie and Huang, Jiancheng and Qiao, Limeng and Ma, Lin},
  journal={arXiv preprint arXiv:2512.13752},
  year={2025}
}

@article{latentum,
  title={LatentUM: Unleashing the Potential of Interleaved Cross-Modal Reasoning via a Latent-Space Unified Model},
  author={Jin, Jiachun and Zhou, Zetong and Yang, Xiao and Zhang, Hao and Liu, Pengfei and Zhu, Jun and Deng, Zhijie},
  journal={arXiv preprint arXiv:2604.02097},
  year={2026}
}

@article{sensenova,
  title={Sensenova-u1: Unifying multimodal understanding and generation with neo-unify architecture},
  author={Diao, Haiwen and Wu, Penghao and Deng, Hanming and Wang, Jiahao and Bai, Shihao and Wu, Silei and Fan, Weichen and Ye, Wenjie and Tong, Wenwen and Fan, Xiangyu and others},
  journal={arXiv preprint arXiv:2605.12500},
  year={2026}
}

@article{ernie_image,
  title={ERNIE-Image Technical Report},
  author={Liu, Jiaxiang and Feng, Zhida and Zou, Pengyu and Qian, Zhenyu and Zhu, Tianrui and Xia, Jun and Dong, Yuehu and Lin, Yanzheng and Xiong, Honglin and Chen, Anqi and others},
  journal={arXiv preprint arXiv:2605.25347},
  year={2026}
}

\end{document}